\documentclass{article}
\usepackage[a4paper, total={6.5in, 10in}]{geometry}
\usepackage{setspace}
\usepackage{graphicx}
\usepackage{multirow}
\usepackage[utf8]{inputenc}
\usepackage[T1]{fontenc}
\usepackage{longtable}
\usepackage[dvipsnames]{xcolor}
\usepackage{amssymb}

\title{From Data to Insights: A Comprehensive Survey on Advanced Applications in Thyroid Cancer Research}

\author{Xinyu Zhang, Vincent CS Lee, Feng Liu}
\date{}

\begin{document}
\onehalfspacing

\maketitle

\begin{abstract}
Thyroid cancer, the most prevalent endocrine cancer, has gained significant global attention due to its impact on public health. Extensive research efforts have been dedicated to leveraging artificial intelligence (AI) methods for the early detection of this disease, aiming to reduce its morbidity rates. However, a comprehensive understanding of the structured organization of research applications in this particular field remains elusive. To address this knowledge gap, we conducted a systematic review and developed a comprehensive taxonomy of machine learning-based applications in thyroid cancer pathogenesis, diagnosis, and prognosis. Our primary objective was to facilitate the research community's ability to stay abreast of technological advancements and potentially lead the emerging trends in this field. This survey presents a coherent literature review framework for interpreting the advanced techniques used in thyroid cancer research. A total of 758 related studies were identified and scrutinized. To the best of our knowledge, this is the first review that provides an in-depth analysis of the various aspects of AI applications employed in the context of thyroid cancer. Furthermore, we highlight key challenges encountered in this domain and propose future research opportunities for those interested in studying the latest trends or exploring less-investigated aspects of thyroid cancer research. By presenting this comprehensive review and taxonomy, we contribute to the existing knowledge in the field, while providing valuable insights for researchers, clinicians, and stakeholders in advancing the understanding and management of this disease.
\end{abstract}

\vskip 2cm

\section{Introduction}

The thyroid is the largest endocrine gland in an adult, shaped like a butterfly and locates at the lower neck. 
It controls the metabolism of cells by producing hormones and regulating the balance of calcium in the human body. 
With an adequately functioned thyroid gland, one can maintain the right amount of hormones to keep metabolism activities at a favorable rate \cite{Salman2019}. 
However, this crucial gland is gradually grabbing attention worldwide since it gives rise to the most common endocrine tumors \cite{Balo2022}.

Thyroid tumors are prevalent and regarded as the most frequently seen nodular lesions among adults. 
Specifically, up to $50\%$ of adults have thyroid nodules \cite{Ach2014}.
$5\%$ to $15\%$ of these nodules turned out to be malignant, forming into four types of thyroid cancer: papillary thyroid carcinoma (PTC), follicular thyroid carcinoma (FTC), medullary thyroid carcinoma (MTC), and anaplastic thyroid carcinoma (ATC) \cite{Nix2005}. 

Since the $20th$ century, thyroid cancer has been in a progressive pattern, rising at the fastest rate among all the malignancies \cite{Lav2015}.
According to the latest Cancer Fact \& Figures statistics, there will be an estimation of $43,800$ new cases diagnosed with thyroid cancer in the United States in $2022$ \cite{ACS2022}.
Thyroid cancer instances are rapidly increasing, with the highest rates found in the Federated States of Micronesia, French Polynesia, North America, and East Asia \cite{Mar2022}.
In particular, South Korea, Cyprus, and Canada had the highest incidence-to-mortality rates \cite{Mar2022}.
In China, thyroid cancer was even ranked as the $4th$ commonly diagnosed disease for females \cite{Feng2019}.

With the apparent implications of increased morbidity and mortality rates brought by thyroid cancer, complications of understanding the cause of the disease, enhancing the diagnostic performance, and offering patients targeted treatments, all mitigating recurrence rates; thus, related existing methodologies should be systematically analyzed.
In this scenario, studies applying statistical and machine learning techniques on investigating the pathogenesis of thyroid cancer, increasing diagnostic accuracy and efficiency, and building patient-centric treatment recommendation systems are expected to be analyzed.
Notwithstanding the potential challenges brought by the long-established machine learning techniques, the advantages of this technology often take place.
Specifically, machine learning-driven techniques adopt various medical tools by taking into account the roles of physicians and computers simultaneously. 
With these techniques, doctors can gain a ``second opinion'' relying on decisions made by computers; also, clinicians get to focus more on patient care, enabling more targeted and customized treatment plans.
Therefore, this survey provides a comprehensive literature analysis regarding applying machine learning techniques for thyroid cancer pathogenesis, diagnosis, and prognosis. 

Following contributions were proposed in this survey paper:
\begin{itemize}
    \item We proposed a comprehensive literature review framework in this study.
    The proposed framework can be adaptable to multi-disciplines for conducting literature analysis.
    
    \item This study offers a structured organization of existing studies regarding machine learning application on thyroid cancer, including its pathogenesis, diagnosis, and prognosis.
    
    \item We provide informative concepts about traditional machine learning techniques and interpret current state-of-the-art deep learning models; this can be considerably helpful for researchers to understand the theory behind deep learning approaches, contributing to more advanced computer-aided diagnosis (CAD) designs for thyroid disease.
    
    \item Current challenges faced by the machine learning approaches will be explained in this study, and corresponding potential research directions will also be outlined.
\end{itemize}

The remainder of the survey is structured as follows: Section $2$ covers the clinical background of thyroid cancer, highlighting the challenges faced by clinicians. 
Section $3$ presents the proposed systematic literature review framework used as the methodology approach. 
Section $4$ depicts the machine learning techniques for thyroid cancer pathogenesis, diagnosis, and prognosis. 
Section $5$ discusses the challenges faced by existing studies in more detail and offers potential suggestions. 
Finally, Section $6$ concludes this survey paper.

\section{Background and Motivation}

Early detection of thyroid cancer leads to mitigated mortality and morbidity rates. 
There exist long-established protocols for the diagnosis and treatment of thyroid cancer in the clinical setting.
Being aware of the pros and cons of those standardized procedures, applications of machine learning-based techniques can be initiated to reap the pros' benefits while overcoming the cons.
We aim to comprehend what, how, and why these machine learning techniques perform in the clinical field.
This section first illustrates the existing clinical protocols for the pathogenesis, diagnosis, and prognosis of thyroid cancer, interprets current challenges faced by clinicians, and then emphasizes the motivations of our study.

\subsection{Thyroid Cancer Pathogenesis}

Although gender is a global consensus factor for thyroid cancer pathogenesis since females have three more possibilities of being diagnosed with thyroid cancer than males \cite{Mau2018}, the risk factors correlated with the disease are still under-researched, much less is to say the interpretability and reliability of the controversial factors. 

\begin{figure}[h]
    \centering
    \includegraphics[width=0.8\columnwidth]{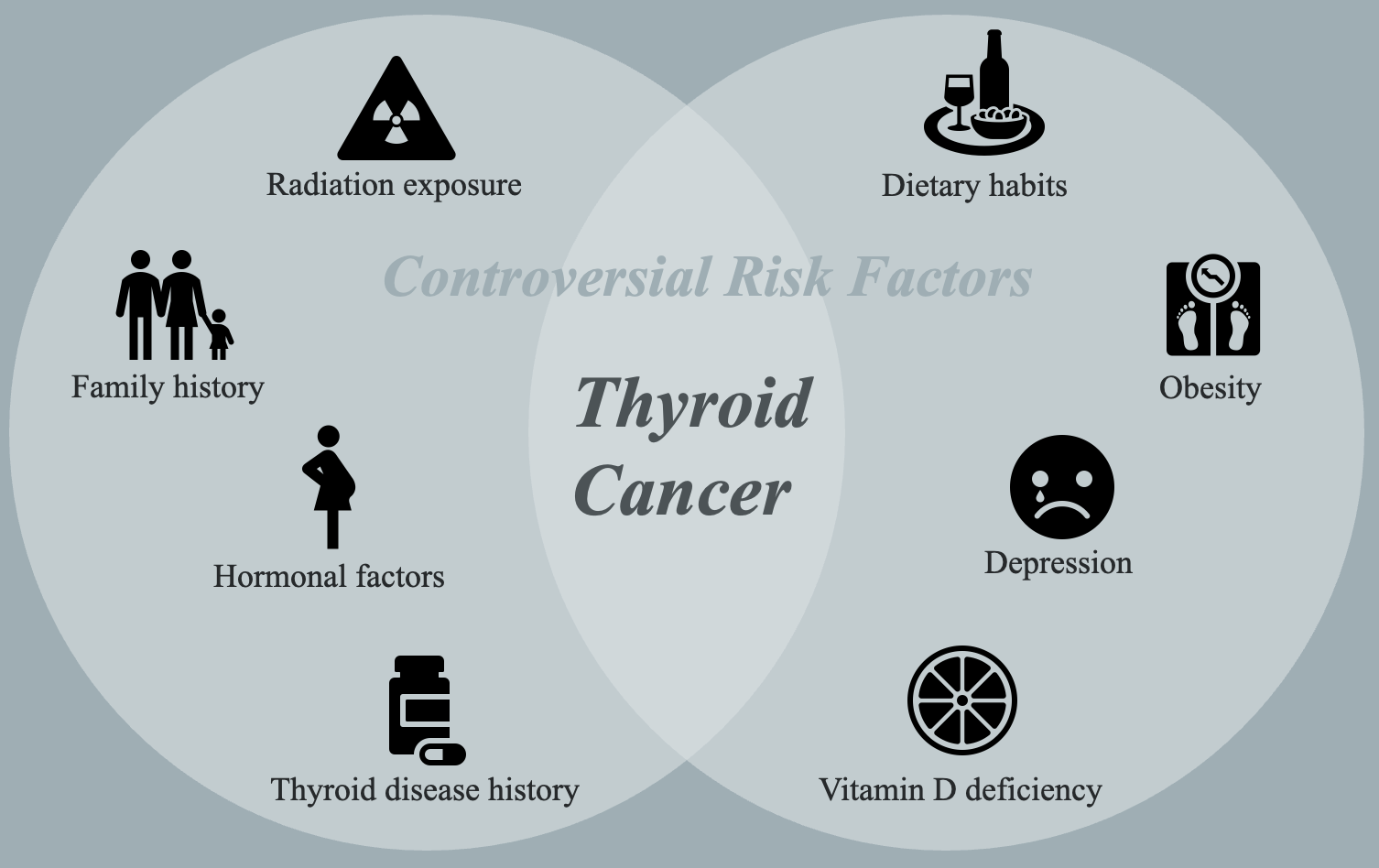}
    \caption{Thyroid cancer controversial risk factors}
    \label{fig:rf}
\end{figure}

Clinicians usually struggle to determine the cause of thyroid cancer, as the pathogenesis behind the disease is affected by multiple diversified attributes (e.g., dietary habits, medical history, gene heredity, and periodical health conditions).
The inherent process is that clinicians always select one factor and evaluate its association with thyroid cancer development.
Following this intrinsic step, several potential factors and comorbidity were identified correlated to thyroid cancer, such as vitamin D deficiency \cite{Min2022}, radiation exposure \cite{Suz2019}, diabetes \cite{Sil2018}, obesity \cite{Jie2015}, iodine intake \cite{Heng2019, Kyu2021}, smoking status \cite{Ara2018}, and family history \cite{Kust2018}, to name a few.
More controversial factors can be found in Figure \ref{fig:rf}.
In this regard, the evaluations of either one of those attributes take long-established retrospective investigations, let alone the interwoven among them; ignoring the correlations among diversified risk factors can bring substantial errors in understanding the cause of thyroid cancer.

To gather the significant points, confirming the associations between a single factor and thyroid cancer development at a time is challenging and inefficient, making thyroid cancer risk factors identification tasks scarcely achievable.
In this case, the associations among multiple factors cannot be evaluated, amplifying the misinterpretations in understanding thyroid cancer cause.
These shortcomings create limitations for revealing the pathogenesis of thyroid cancer; therefore, we urgently need the means to incorporate an automated algorithm to efficiently mine essential knowledge from high-dimensional medical records to reveal the pathogenesis of thyroid cancer.

\subsection{Thyroid Cancer Diagnosis}

Following the clinical guidance, the thyroid function examination is always in diagnostic priority, which measures hormones produced by the gland. 
It includes measuring thyroid-stimulating hormone (TSH), triiodothyronine (T$3$), thyroxine (T$4$), free triiodothyronine (FT$3$), and free thyroxine (FT$4$) \cite{Kline2017}.
The patient will be suggested an ultrasonography test when identifying abnormal thyroid function results. 
Ultrasonography provides an intermediate risk following the \emph{Thyroid Imaging Reporting and Data System} (TIRADS) score based on the features appearing on the ultrasound images.
TIRADS measures an image based on nodule features like irregular margins, hypo-echogenicity, taller-than-wide shape, and microcalcifications \cite{Russ2011}. 
TIRADS scores rank from $1$ to $5$ where $1$ was defined as ``normal thyroid'', $2$ as ``benign features'', $3$ as ``absence of suspicious features'', and $4a$ as ``one suspicious feature’’, $4b$ as ``two suspicious features'', $4c$ as ``three to four suspicious features'', and $5$ as ``five or more suspicious features'', respectively.

Nevertheless, TIRADS scores cannot provide precise decisions for thyroid cancer diagnosis.
Based on the analysis conducted by \cite{ATA2018} in which $951$ patients' TIRADS scores were compared to the ground-truth histopathological results.
The authors found that thyroid cancer was appearing at a $0.9\%$ with TIRADS ranking $2$, $2.9$\% cancer rate was found in TIRADS $3$, TIRADS $4a$ had a cancer rate of $12.3$\%, $34.4$\% cancer rate was found in TIRADS $4b$, $66.6$\% in TIRADS $4c$, and $86$\% cancer rate was in TIRADS $5$. 
Driven from this finding, solely relying on ultrasound images to make a diagnosis through TIRADS scores is considered insufficient. 
In this scenario, patients having TIRADS scored from $2$ to $5$ will need to perform further fine-needle aspirations cytology (FNAC) to determine whether the nodule has cancerous cells \cite{Stew2020}.

\begin{figure}[h]
    \centering
    \includegraphics[width=0.7\columnwidth]{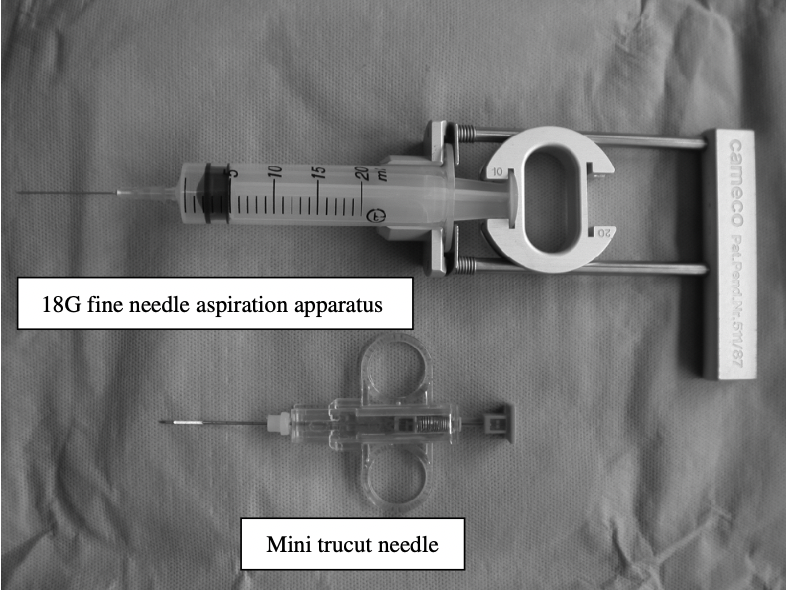}
    \caption{FNA Apparatus (samples from \cite{Nix2005})}
    \label{fig:tool}
\end{figure}

\begin{figure}
    \centering
    \includegraphics[width=0.7\columnwidth]{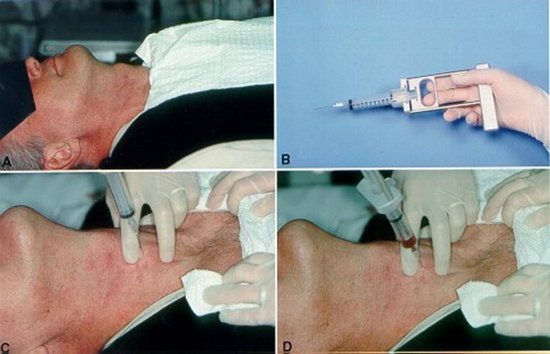}
    \caption{FNA Procedures (samples from \cite{Dean2015})}
    \label{fig:fna}
\end{figure}

FNAC is regarded as the golden standard in assessing malignancy of thyroid nodules \cite{Baloch2002}. 
It gets a biopsy from nodule cells using needles following the ultrasonography guidance, then has a biopsy assessed by pathologists to make diagnostic decisions regarding nodule malignancy risk stratification. 
Figure \ref{fig:tool} demonstrates the FNAC apparatus and Figure \ref{fig:fna} displays the implementation procedures.
However, FNAC examinations are easily affected by the professionalism and expertise of pathologists, resulting in $30$\% of FNAC results being non-diagnostic or indeterminate \cite{Ouyang2019}. 
Under this circumstance, patients with indeterminate FNAC results usually need to undergo another FNAC or even an excisional biopsy for further diagnostic decision-making.
Sometimes patients might undergo unnecessary surgeries, then determining the nodule shows no evidence of malignancy. 

These shortcomings create limitations for thyroid cancer diagnosis in the clinical setting.
Therefore, clinicians seek to improve diagnostic accuracy and efficiency by integrating human knowledge with computer-driven techniques. 
With the emergence of machine learning, particularly the deep learning concept, the diagnostic procedures form into advanced ways of making decisions, naming ``computer-aided diagnosis (CAD)'' techniques.

\subsection{Thyroid Cancer Prognosis}

Thyroid cancer has a rigorous treatment protocol.
Specifically, patients diagnosed with thyroid cancer, regardless of the subtype, require thyroid surgery (i.e., partial thyroidectomy, total thyroidectomy, or total thyroidectomy with lymph node dissection).
Further radiation therapy is expected to mitigate recurrence risks by performing Iodine-$131$ treatment.
Patients with thyroidectomy are expected to take levothyroxine one hour before breakfast for thyroid hormones replacements \cite{Walsh2016}. 
Levothyroxine is a manufactured form of the thyroid hormone thyroxine, and the common thyroxine medications are Eutroxsig, Oroxine, and Aspen Pharma \cite{Walsh2016}. 
 
Noteworthy, over-treatments are substantially occurred with thyroid disease \cite{Jeg2017}. 
The post-therapy hypothyroid disease frequently occurs after surgery, or Iodine-$131$ treatments \cite{Erik2002}. 
Radio-iodine therapy would somehow bring adverse effects for patients \cite{Schmid2017}.
\cite{Lee2015} also confirmed that Iodine-$131$ treatment would cause salivary gland dysfunction if the dosage of iodine were not carefully determined. 
Thus, over-treatment must be avoided by offering patients customized treatment plans based on their age, weight, and cancer stage, to name a few. 
However, there is a scarcity of studies that propose customized treatment recommendation systems for patients with thyroid cancer.

Besides providing personalized treatment plans, predicting the recurrence and survival rates can also potentially enhance the prognosis for thyroid cancer patients.
An estimation of $2,230$ people will die from thyroid cancer in $2022$ \cite{ACS2022} with a mortality rate of $5\%$.
Among the four types of thyroid cancer, PTC is the most commonly diagnosed kind with the highest survival rate.
Specifically, the five-year survival rate of PTC is more than $98\%$ \cite{Lav2015}; patients with PTC are usually expected to have a normal lifespan after surgery and radiotherapy treatments. 
On the contrary, rarer kinds like anaplastic or medullary have much lower survival rates, which is less than $10\%$ \cite{Melis2020}.

With respect to the recurrence of thyroid cancer, \cite{Maz1994} indicated that $20\%$ of patients would experience the recurrence, in which they might need re-operations or repetitive radiotherapy treatments. 
Unlike the pathogenesis of thyroid cancer, the risk factors correlated to thyroid cancer recurrences are not too controversial, including gender, elder age, primary disease extent, metastases to other organs, tumor size, extra-thyroidal invasion, location of nodules, and cervical lymph nodules are the leading factors for thyroid cancer recurrence \cite{Simps1987, Bors2004, Pel2007, Ray2013, KimS2014, Mourad2020}. 
\cite{WangT2013} proposed that PTC patients who had a total thyroidectomy and neck dissections tend to have a $3.8\%$ recurrence rate. 
In addition, different surgical treatments also differentiate recurrence rates of thyroid cancer. 
For instance, based on the comparative study by \cite{Zet2010}, the recurrence rate for patients with thyroidectomy and neck dissection is $2.02\%$, whereas the recurrence rate for patients with thyroidectomy only is $3.92\%$. 
It is evident that thyroidectomy and iodine treatment decreases the recurrence rate of thyroid cancer \cite{Maz1994, Pel2007, Van2009}. 

Collectively, thyroid cancer prognosis is closely related to the well-established treatment protocols.
Existing studies harnessed the power of data mining to extract valuable knowledge from thyroid cancer patients in helping provide personalized treatment plans and make predictions in recurrence to achieve a patient-oriented treatment recommendation system so that patients and experts can access self-driven decision-making to optimize the prognosis of thyroid cancer.

\section{Literature Review Methodology}

\begin{figure}
    \centering
    \includegraphics[width=0.78\columnwidth]{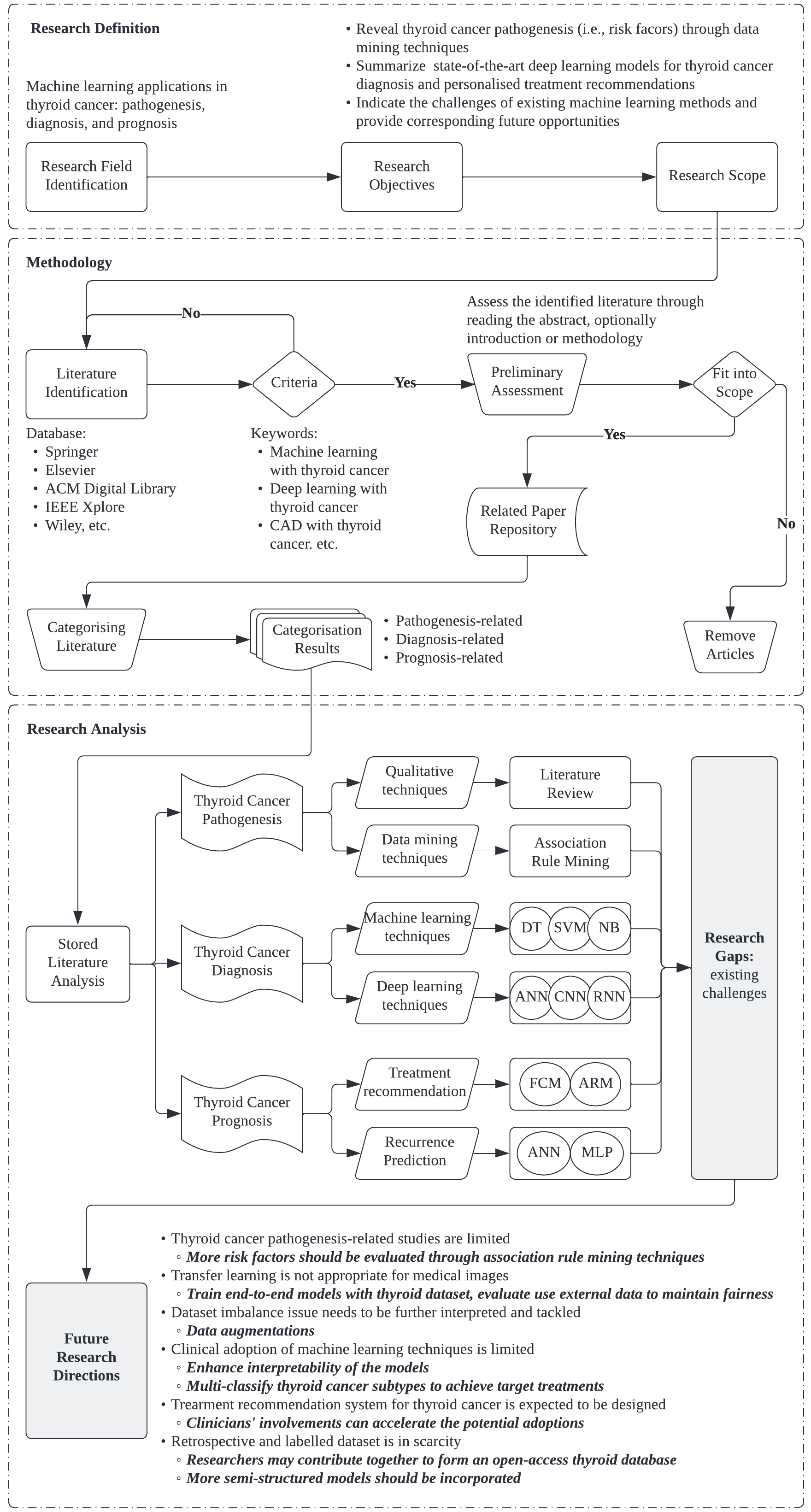}
    \caption{Systematic literature review framework}
    \label{fig:framework}
\end{figure}

In order to conduct a comprehensive systematic literature analysis, this paper proposes a detailed step-by-step literature review framework presented in Figure \ref{fig:framework}.

The overall literature review framework contains three stages: research definition, research methodology, and research analysis. 
``Research definition'' includes identifying the research field, defining research objectives, and outlining the research scope. 
``Research methodology'' stage oughts to identify all the related literature studies based on the pre-defined searching protocols. 
``Research analysis'' stage then analyses the identified studies, demonstrating the critical findings and explaining the existing literature gaps while interpreting future research opportunities. 
In this section, the three stages will be explained in detail. 

\subsection{Research Definition}

The first stage of the literature review framework is the research definition. 
The research focus has been identified in this phase; meanwhile, the research motivations and objectives will need to be highlighted, and the research scope will be defined.

In this research, we focus on understanding machine learning-based applications for thyroid cancer, not only for the diagnosis but also to embrace the studies conducted around thyroid cancer risk factors, treatments, and prognosis. 
Existing literature review studies simply investigate machine learning approaches for thyroid disease detection, ignoring the advanced algorithms results in the obsolete knowledge discovery, much less is to summarise the findings from the pathogenesis and prognosis facets. 
While the early detection of thyroid disease can certainly reduce morbidity and mortality rates, the pathogenesis and prognosis are as necessary as the diagnosis to achieve a well-established prognosis. 
Hence, we ought to include studies from all three perspectives so that the researchers can gain in-depth knowledge to develop more sophisticated models. 

We seek to identify papers that incorporate machine learning algorithms to help reveal thyroid cancer pathogenesis, achieve advanced CAD design to release patients from painful clinical diagnostic procedures, and design a patient-centric treatment recommendation system to achieve a promising prognosis. 
Therefore, we aim to analyse literature that aims to comprehend, diagnose, and treat thyroid cancer through data mining, machine learning, or deep learning techniques. 

Lastly, the scope of this survey paper is to include as much related literature as possible for analysis. 
Machine learning techniques have been applied quite often in the medical field, yet the application to thyroid cancer is relatively limited. 
In order to have a comprehensive analysis, we intend to include as many high-level ranking literature studies as possible. 
Moreover, the specific literature identification process will be explained next.

\subsection{Research Methodology}

After defining the research scope, a set of literature identification processes was conducted. 
We have included a list of high-level ranking conference and journal databases during the literature searching phase, such as Elsevier, Springer, IEEE Xplore, and ACM Digital Library, to name a few. 
A set of criteria were used as our search protocols, and Table \ref{tab:1} lists some keywords applied.

\begin{table}[h]
    \centering
    {\bf \caption{Literature identification searching phase}}
    \begin{tabular}{p{6cm}l}
        \hline
        {\bf Research Focus} &  {\bf Searching Keyword Phases} \\
        \hline
        \multirow{4}{*}{Thyroid cancer pathogenesis} & Risk factors of thyroid cancer \\
        & Thyroid cancer pathogenesis with machine learning \\
        & Thyroid cancer pathogenesis with data mining \\
        & Thyroid cancer risk factors with association rule mining \\
        \hline
        \multirow{3}{*}{Thyroid cancer diagnosis} & Machine learning with thyroid cancer \\
        & Deep learning with thyroid cancer \\
        & CAD for thyroid cancer detection \\
        \hline
        \multirow{3}{*}{Thyroid cancer prognosis} & Thyroid cancer treatment recommendation system \\
        & Machine learning with thyroid cancer treatment \\
        & Prediction of thyroid cancer survival or recurrence \\
        \hline
    \end{tabular}
    \label{tab:1}
\end{table}

\begin{figure}[h]
    \centering
    \includegraphics[width=0.7\columnwidth]{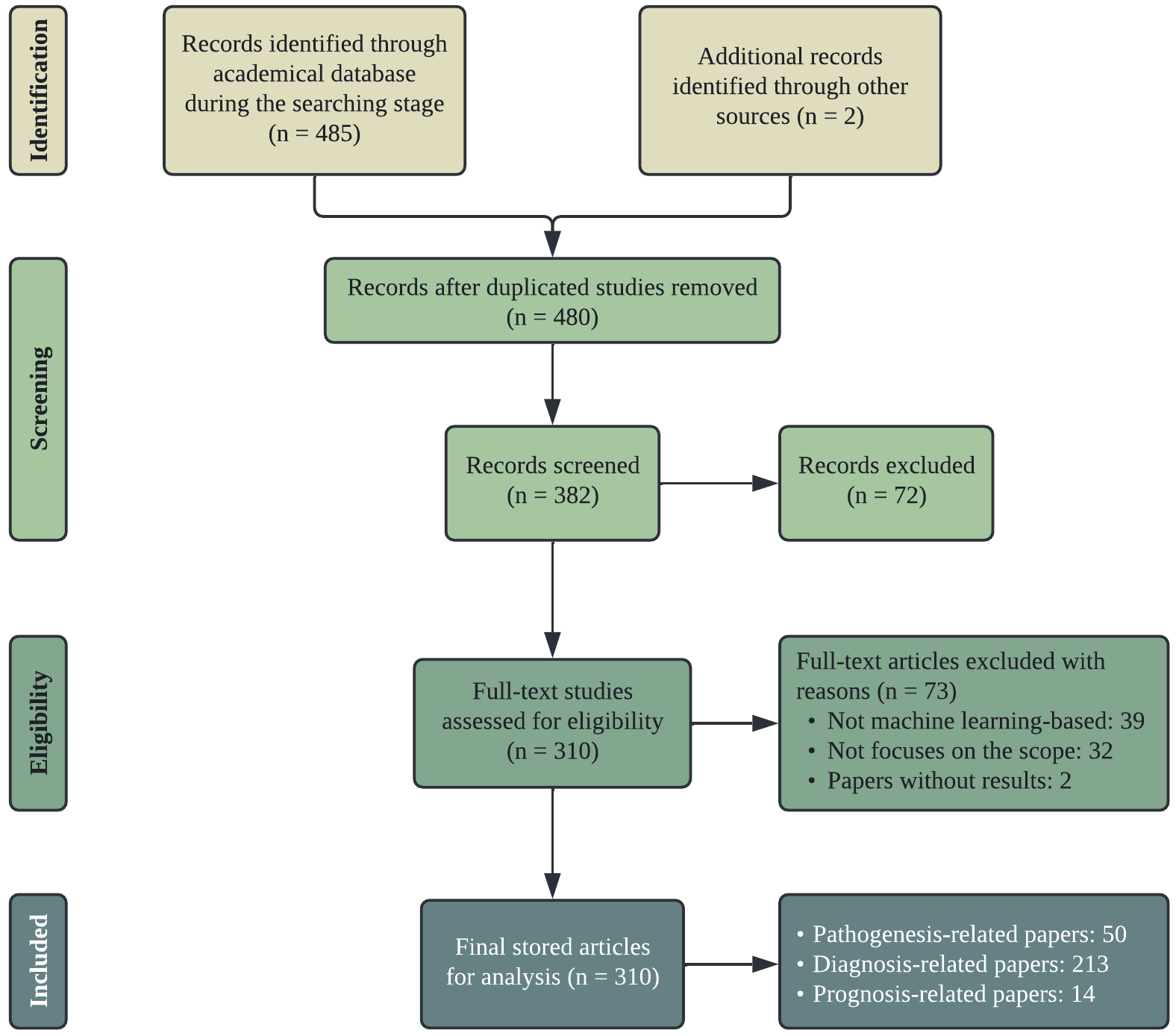}
    \caption{Flowchart of the searching strategy and literature selection}
    \label{fig:5}
\end{figure}

Since deep learning techniques are relatively advanced, we want to explore the developments of these approaches over time; thus, during the searching phase, no timeline restriction was followed. 
In this regard, the literature studies were selected based on the flowchart (Figure \ref{fig:5}) when applying the above-identified keywords in the academic database. 

\begin{figure}[h]
    \centering
    \includegraphics[width=0.9\columnwidth]{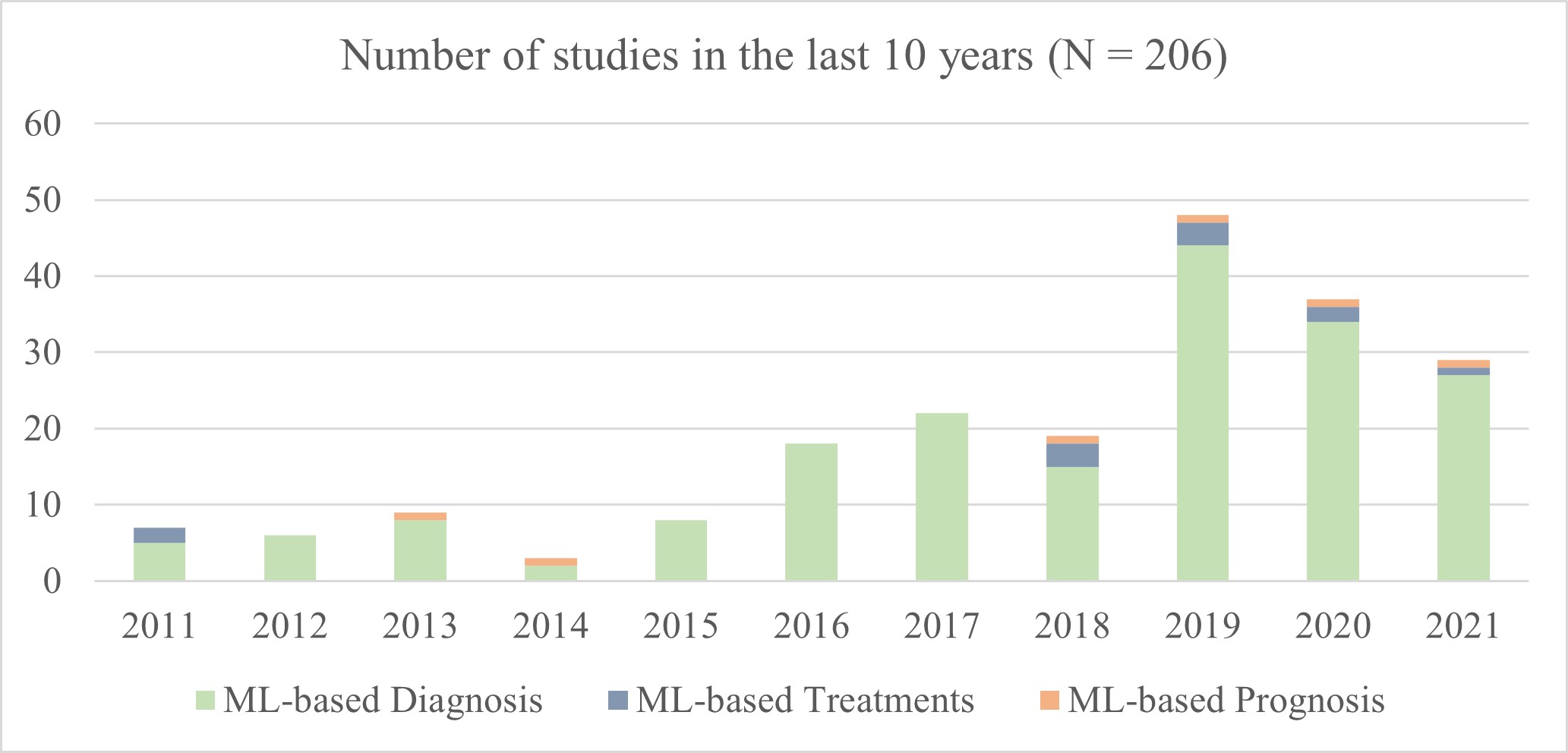}
    \caption{Number of literature distribution in the last $10$ years}
    \label{fig:number}
\end{figure}

Precisely, we have carefully read through the abstract of each paper.
Optionally we would read through the methodology section if the abstract were not explicit enough regarding whether machine learning techniques were used. 
Then, the papers that fit into our pre-defined scope will be stored in the literature repository for further analysis, otherwise removed. 
Lastly, all the identified papers were categorized into three groups: pathogenesis, diagnosis, and prognosis. 
As a result, $70$ papers are categorized into the pathogenesis group, $213$ papers are categorized into the diagnosis group, $24$ papers are about the prognosis of thyroid cancer. 
Details of the paper distribution can be viewed in Figure \ref{fig:number}.

Figure \ref{fig:number} exhibits that the number of papers on thyroid disease diagnosis is way much higher than the pathogenesis and prognosis-related studies.
Furthermore, most related studies were published in $2019$, while the least number of papers was published in $2014$. 
Since thyroid disease is rapidly rising, more and more research studies have been conducted in the last $10$ years; thus, a gradual increase can be found in the figure.

\subsection{Research Analysis}

The research analysis is the last stage of the systematic literature review approach. 
A total number of $136$ papers were utilised for the analysis phase in this survey paper. 
Each of the three categories was analysed based on the performance, feasibility, and generalizability of the proposed machine learning approaches. 
To better interpret our findings, the analysis section was categorised into three main perspectives aligning with our research scope.

For thyroid cancer pathogenesis-related studies, we have divided the literature into qualitative and quantitative groups. 
Qualitative studies investigated risk factors correlated with thyroid cancer, and quantitative studies further confirmed some risk factors through data mining techniques.
For thyroid cancer diagnosis-related studies, machine learning and deep learning applications were discussed independently. 
As for thyroid cancer prognosis-related studies, the design of treatment recommendation systems and the prediction of survival, death, and recurrence rates were explained. 
At the end of each section, existing literature gaps and currently facing challenges will be outlined. 
Collectively, a summary of the literature analysis will be presented, existing challenges inferred from literature analysis will be outlined, and corresponding future research directions will be provided for researchers.

\section{Machine Learning Applications in Thyroid Cancer}

With the emergence of machine learning techniques in $1960s$ \cite{Igor2001}, their applications have been implemented to enable modelling and analyzing large scales of datasets.
Various methods have been deployed in the medical domain to address clinical dilemmas.
A summary of major machine learning techniques around thyroid cancer applications is provided in this section.

\subsection{Applications in Thyroid Cancer Pathogenesis}

Thyroid cancer pathogenesis has always been on hit in the clinical and academic domains. 
Qualitative and quantitative analysis was usually conducted for revealing such a mystery.
Well established factors like radiation exposure in childhood \cite{Atha2019} and gene heredity \cite{Kinga2019} are in consensus; whereas factors like diabetes \cite{Yeo2014}, obesity \cite{Danu2018}, vitamin D deficiency \cite{Luigi2021}, hormonal factors \cite{Lulu2022}, are still under debate.

With the identified studies, qualitative literature was based on survey analysis, whereas quantitative studies were majorly built based on statistical learning, including case-control, retrospective, or prospective cohort studies.
The machine learning application in the clinical domain is considerably limited with respect to thyroid cancer pathogenesis identification, leading to insufficient work produced.
In this regard, this section interprets the commonly utilised machine learning algorithms.

Among the identified machine learning-related studies, data mining, specifically, association rule mining (ARM) techniques have been applied relatively often.
ARM is dedicated to revealing hidden patterns among complex, high-dimension, and large volumes of interwoven attributes \cite{Mina2021}; its applications in the clinical domain demonstrate superior performance in comorbidity detection \cite{WangChi2019, ChaS2021}, disease symptom analysis \cite{Meer2021}, and behavioural analysis for driving certain disease \cite{LeeS2021}.

ARM techniques generate association rules denoted as $X \Rightarrow Y$, in which $X$ is the antecedent item and $Y$ is the consequence, meaning when $X$ exists in an observed instance, then item $Y$ should co-exist.
The high-quality rules are assessed through the support and the confidence metrics.
The support value can be seen as the frequency of the itemset that appeared in the database, calculated through Eq. \ref{eq:sup}, where $N$ is the total number of instances in the database.
The confidence is a conditional probability that determines the occurrence of the itemset $X$ and $Y$ given the probability of antecedent $X$ itself, calculated through Eq. \ref{eq:con}.
Besides the support and confidence metrics, strongly-related ARM rules sometimes are also selected based on a lift measurement.
The lift value suggests the ratio of the probability of having $X$ and $Y$ simultaneously compared to the probability of having $X$ and $Y$ independently, which is calculated through Eq. \ref{eq:lift}.
Specifically, when the lift value is high, it is more likely that the co-existence of $X$ and $Y$ is not just a random occurrence \cite{Buc2010}.

\begin{equation}\label{eq:sup}
    Support = \frac{freq(X \Rightarrow Y)}{N}
\end{equation}

\begin{equation}\label{eq:con}
    Confidence = \frac{freq(X \Rightarrow Y)}{freq(X)}
\end{equation}

\begin{equation}{\label{eq:lift}}
    Lift = \frac{Support(X \Rightarrow Y)}{P(X) \times P(Y)}
\end{equation}

With those fundamental metrics, extensive studies mined valuable information from heterogeneous database for knowledge discovery.
Although clinical research applying ARM is rich, the adoption in identifying thyroid disease pathogenesis is considerably limited, with only two pieces of literature were found amid.
\cite{Dong2020} once deployed the Apriori and the Predictive Apriori algorithms to find risk factors correlated with thyroid disease, and further input the strongly related factors into a back-propagation neural network for binary classification between healthy and sick groups. 
A similar study was conducted by \cite{Dongy2020}, in which the Apriori algorithm was incorporated with the bagging algorithm for thyroid disease detection.

\subsubsection{Apriori}
The Apriori algorithm was proposed by \cite{Agra1994} in $1994$, aimed to discover underlying association rules, also known as frequent itemsets, given a large transaction database.
The algorithm was built as tabular-based and required the generation of candidate itemsets through iteration. 
In the first iteration, the occurrence of each candidate item $k$ would be counted and recorded, whereas the rest of the items below the minimum threshold would be pruned.
In this regard, the frequency of the item is known as the support value; thus, the minimum threshold would be the minimum support value defined prior to the implementation of the algorithm.
In the second iteration, an increment $(k+1)$ itemsets would again be assessed with the threshold.
The algorithm scans through the entire database in each iteration until it finishes counting the occurrence of all possible combinations for the candidate itemsets.
To this extent, the Apriori algorithm was considered as the most classic ARM technique as it is easy to be implemented and understood.  

Nevertheless, the major shortage of the Apriori algorithm is with its efficiency.
The algorithm requires the generation of candidate itemsets and scanning through the entire database in each iteration, this limits the running time of the algorithm.
In particular, medical datasets are vast in volume and complex in dimension, making the workload of the Apriori algorithm even more cumbersome.
Although considered as accurate in determining frequent itemsets, the Apriori algorithm is not computationally efficient enough when the dataset gets more extensive or complicated.

\subsubsection{Predictive Apriori}

The generated association rules can be classified into common rules (i.e., rules with high support and confidence), reference rules (i.e., rules with low support and confidence), and exception rules (i.e., rules with low support but high confidence) \cite{Dav2008}.
Common rules are the frequent itemsets which we aim to identify through ARM techniques.
Reference rules are usually discarded as they merely produce valuable knowledge.
Exception rules, on the other hand, should not be overlooked since they are beneficial and seen as actionability for policy establishment, system development, decision-making, and risk factors prevention.
The Apriori algorithm requires the selection of support and confidence thresholds, this process is risky as exception rules can easily be pruned out when we set the support threshold too high.
The Predictive Apriori algorithm was introduced by \cite{Tobias2001}, made up to the limitation and dedicated to mitigating the risks of trimming valuable exception rules.

Compare to the Apriori algorithm, the Predictive Apriori algorithm does not require configuring thresholds for support and confidence metrics.
Instead, the algorithm determines an expected number of rules that want to be obtained based on the ``predictive accuracy''.
In such a process, the algorithm can prune redundant association rules efficiently.
However, the pre-defining of the predictive accuracy and the limitation of efficiency are still presented for the Predictive Algorithm, as it also requires scanning through the entire database continuously.

With the ARM techniques, the most strongly-related factor with thyroid disease is elder age.
The age group ranging from $60$ to $80$ is very likely to be diagnosed with thyroid diseases, such as hypothyroidism or hyperthyroidism.
The age factor has always been discussed in the field.
\cite{Mimi2021} once deployed qualitative analysis for thyroid cancer risk factors investigation, and they found that middle-aged people, specifically those aged from $50$ to $54$ are at high risk of being diagnosed with thyroid cancer compared to other age groups.
Contradictory studies are often presented.
A case-control study was conducted by \cite{Mag2018}, and their results demonstrated that younger age is statistically significantly related to the risks of establishing thyroid cancer.
The retrospective study implemented by \cite{Azi2011} also indicated that thyroid cancer is likely to be established in people age younger than $55$. 

Other qualitative studies found risk factors correlated with the increasing rate of thyroid cancer, including the history of thyroid disease \cite{Papan2019, Atha2019}, obesity \cite{Sad2017, Mat2020, Carim2020}, iodine deficiency \cite{Zimm2015}, dietary habits \cite{Wook2014}, family history \cite{Atha2019}, diabetes \cite{Yeo2014, Luigi2021}, hormonal factors \cite{Mar2014}, and stress \cite{Kyr2022}, to name a few. 
Through the evaluation of quantitative methodologies, such as logistic regression and cox regression, confirmed factors include the history of thyroid disease (i.e., Hashimoto's thyroiditis, thyroid nodules) \cite{Vlad2013} and family history \cite{Asif2015}. 
However, the remaining factors are still controversial and cannot be ascertained based on existing studies, and these factors are obesity, iodine deficiency, dietary habits, diabetes, and hormonal factors.
\cite{Kyle2015} once established a retrospective study with $467$ patients examined through case-control technique, and they found tobacco use, alcohol consumption, and obesity had no association with the increasing risk of thyroid cancer; this finding aligns with \cite{Danu2018}.
Additionally, iodine intake is always a debating factor, yet whether its excessive or deficient consumption will lead to thyroid cancer is still unknown \cite{Cao2017}.
Similarly, food consumption in daily dietary \cite{Wook2014}, diabetes \cite{Chin2012}, and hormonal factors \cite{Mein2009} in relation to thyroid cancer risk tend to be weak and inconsistent.

\subsubsection{Thyroid Cancer Pathogenesis Literature Summary}

\begin{table}[h]
\label{tab:risk}
    \centering
    {\bf \caption{Thyroid Cancer Risk Factors (Last $10$ Years)}}
    \begin{tabular}{p{4.8cm}p{4cm}p{7cm}}
        \hline
        {\bf Reference} & {\bf Methodology} & {\bf Findings} \\
        \hline
        \cite{Yul2013} & Multi-variable analysis & Presence of Hashimoto's thyroiditis, Autoimmunity \\
        \cite{Vlad2013} & Logistic regression (LR) & Smoking, History of thyroid nodules, History of non-thyroid cancer, Diabetes, Radiotherapy of head or neck \\
        \cite{Wook2014} & Survey & Consumption of meat, Regular use of multivitamins, Dietary nitrate and nitrite \\
        \cite{Yeo2014} & Survey & Diabetes \\
        \cite{Mar2014} & Survey & Thyroid hormones, Insulin resistance, Adipokines, Inflammation, Sexual hormones \\
        \cite{Xha2014} & Case-control study & Nuclear fallout before $15$, History of medical radiation, Obesity, Tallness, Artificial menopause, Family history, Iodine deficiency, Spring drinking water \\
        \cite{Zimm2015} & Survey & Iodine deficiency \\
        \cite{Asif2015} & LR & Marital status, Family history, Dietary iodine, Oxidative stress, Fast and fried food \\
        \cite{Cao2015} & Survey & Older age at menopause and parity \\
        \cite{Xha2015} & Case-control study & High body surface area, Great height, Excess weight, High body of fat percentage \\
        \cite{Ping2015} & Survey & Late age at menopause \\
        \cite{Luo2016} & Cohort study & Hysterectomy \\
        \cite{Sad2017} & Survey & Obesity, Overweight, Radiation exposure \\
        \cite{Khod2018} & Survey & Radiation, Smoking, Alcohol, Nutrition elements \\
        \cite{Kar2018} & LR & Obesity, Family history, Use of levothyroxine, TSH \\
        \cite{Mile2018} & Retrospective study & Younger age, Smaller nodule size, Hypo-echoic nodule, Calcification \\
        \cite{Atha2019} & Survey & Radiation exposure during childhood, Family history, Presence of Hashimoto's thyroiditis \\
        \cite{Kinga2019} & Survey & Genetic factors \\
        \cite{Ara2019} & Case-control study & Hormonal factors \\
        \cite{Yuji2018} & Survey & Hashimoto's thyroiditis, Elevated TSH \\
        \cite{Park2020} & Follow-up study & Metabolic syndrome, Obesity \\
        \cite{Mimi2021} & Survey & Geographical factors, Age, Higher BMI \\
        \cite{Luigi2021} & Survey & Iodine deficiency, Diabetes, Pollutants, Radiation \\
        \cite{Lulu2022} & Mendelian randomization & Telomere length, Waist-to-hip ratio, Diastolic blood pressure, Central obesity \\
        \cite{Kyr2022} & Survey & Stress \\
        \cite{HanJ2013, Sitong2019} & Retrospective Study, Case-control study & Body Mass Index (BMI), Obesity \\
        \cite{Nour2015, Kita2018, Papan2019} & Survey, Cohort study & History of thyroid disease \\
        \cite{Shen2018, Dong2020, Dongy2020} & ARM, Follow-up study & Age \\
        \cite{Pelle2013, Brise2014, Yihao2017, Efan2017, Miah2018} & Survey, LR & Radiation exposure \\
        \cite{Pappa2014, Jie2015, Kwon2019, Mat2020, Soo2020, Carim2020} & Survey, LR, Cohort study & Overweight, Obesity \\
        \hline
    \end{tabular}
\end{table}

Prior prospective studies determined that environmental or medical radiation exposure during childhood, genetic factors, history of thyroid disease, and family history are of high risks in relation to the development of thyroid cancer; other factors beyond them need further interpretations and observations (details see Table \ref{tab:risk}).

With the continuous progress of machine learning techniques, the utilisation of ARM algorithms demonstrates increased efficiency in revealing the hidden patterns for thyroid cancer pathogenesis.
Nevertheless, the most significant limitation of classic ARM algorithms is that they can only handle categorical attributes.
In this regard, the presence of a particular symptom or comorbidity can be assessed, whereas further numerical attributes cannot be taken into consideration.
For instance, factors like blood pressure, BMI, or weight has to be discretised and assigned to a definite genre for evaluations. 
Additionally, with the increased complexity of electronic health records (EHR), medical data is with high complexity and dimensionality; thus, the traditional ARM techniques cannot deal with EHR records efficiently and comprehensively.
Another limitation of traditional ARM techniques is the minimum threshold selection for support and confidence, which is still challenging and risky.
These shortcomings create limitations for the ARM applications in the clinical domain.
Ignoring those shortcomings will bring substantial errors or perennial pauses in machine learning applications in the clinical domain.
Accordingly, more advanced ARM techniques are in demand for revealing intricate disease epidemiology.

In this respect, the Frequent Pattern-Growth Tree (FP-Tree) algorithm introduced by \cite{Han2000} harnessed the power of tree structures and enhanced efficiency for mining association rules compared to Apriori.
\cite{Buc2010} proposed fuzzy ARM to incorporate numerical and categorical attributes for findings association rules.
This concept has also been deployed already in the clinical for heart disease diagnosis \cite{Umas2018}.
\cite{TaoF2003} introduced weighted ARM by generating weighted support values for selecting rules, which has also been used for comorbidity detection \cite{LaksV2019}. 
A prior study conducted by \cite{Yav2021} proposed a profile-based ARM technique which generates profiles for the individual patient based on their medical conditions and aims to assess their risks of developing heart disease through the incorporation of fuzzy ARM.
These studies are in the practice of developing precise association rules.
Nevertheless, studies incorporating such ARM techniques are still absent for thyroid cancer pathogenesis identification.
Bridging such a gap will potentially be a breakthrough in the clinical domain.
Moreover, developing a robust, adaptive, and efficient ARM technique which considers the comprehensive attribute formats is also in great demand.

\subsection{Applications in Thyroid Cancer Diagnosis}

CAD applications usually consist of four parts: pre-processing, segmentation, feature extraction, and classification \cite{Anand2020}. 
Medical images acquired from cross-institutions are not standardised due to the differences in devices, policies, or expertise. 
Thus, pre-processing of images is required to remove noises, enhance image quality (i.e., contrast, colours, and sharpness), or even augment dataset to make them consistent and adequate for further analysis \cite{Chan2021}. 
Afterwards, a segmentation step, which extracts the object for detection from the background, is usually required to determine the region of interest (ROI) \cite{Jin2017}. 
Feature extraction primarily selects essential features from the ROI based on domain expert knowledge, and this allows to form into a feature set ready to be fed into the classifier for decision-making \cite{Anand2020}. 
Classification is always the ultimate goal which decides the class of the object (e.g., benign or malignant, stage of a particular disease) based on the input feature sets \cite{Anand2020}. 
Compared to manual diagnosis, CAD makes diagnosis more accurate and efficient. 
CAD mitigates human false-positive rates and achieves diagnosis through computational power, allowing clinicians to focus more on patient care. 
Deriving from machine learning techniques allows CAD applications to become relatively engaging in thyroid disease diagnosis in the era of artificial intelligence.

\subsubsection{Pre-processing}

The essence of machine learning techniques is to train computers to learn from data through statistical analysis; deriving from this, the quality of input data is highly influential to machine learning-based CAD systems.
With the objective to enhance the CAD performance, the input data quality is the primary to be ensured.
Nevertheless, CAD applications generally require cross-institutional evaluations, and the data quality can be affected by varied institutional guidelines or policies.
Under this scenario, standardizing the input data is demanding to enhance its quality.
Besides, data quality can also be promoted based on volume, consistency, and timeliness.
Accordingly, this section presents some data pre-processing techniques for CAD design in the thyroid cancer detection task.

Machine learning techniques make predictions or decisions based on the learnt experience.
To this end, the experience denotes the input data; thus, the more volume of the input data is, the more accurate the predictions will be.
In this regard, many data augmentation techniques were proposed in CAD design.
For instance, \cite{Chan2021} once deployed ultrasound images for detecting thyroid cancer.
The originally acquired dataset consists of $1791$ images, then the authors applied image augmentation techniques, including adjusting image contrasting and horizontal flipping.
Afterwards, the authors augmented the image set to $7360$ ultrasound images for training and testing.
Additionally, some researchers choose to mirror the images for augmentation purposes.
For example, \cite{Bila2020} adopted an open-access ultrasound image set for thyroid nodules recognition, and they have eventually augmented to $4000$ images, whereas the original dataset only contains $451$ images.
Other studies choose to crop images into patches to increase the size of the input data \cite{ZhaoZ2021}, rotate images \cite{Olu2022}, or even add or adjust the Gaussian noises for image augmentation \cite{Shen2020}.
With varied extend of augmentation techniques, the CAD performance can be increased, becoming an indispensable pre-processing step.

\begin{figure}[h]
    \centering
    \includegraphics[width=0.6\columnwidth]{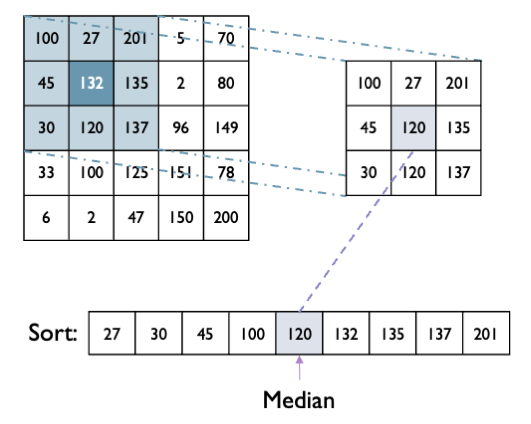}
    \caption{Adaptive median filtering flowchart for noise removal}
    \label{fig:amf}
\end{figure}

In the design of CAD applications for thyroid cancer, the use of medical images is more prosperous than clinical features.
However, the acquired medical images are different in quality based on cross-institutional guidelines, devices, or even the clinician's expertise.
In this regard, standardizing the medical images is necessary for the pre-processing step.
The common tasks are removing image annotations \cite{Hanu2021} and speckle noises \cite{Vad2021}.
For instance, \cite{Chi2017} once deployed the technique proposed by \cite{Nikh2013} to remove artefacts in thyroid ultrasound images.
Precisely, the authors extract the non-zero region from the input image, plot a histogram containing the artefacts, identify the histogram peaks as the intensity level of the artefacts, and subtract the artefact pixels with the intensity levels to restore the image without any annotations \cite{Chi2017}.
Besides, some researchers adopted the adaptive median filtering (AMF) technique for annotations, markers, and noises removal \cite{Hat2020}.
The median filtering algorithm detects the impulse noise by comparing each pixel to its neighbours; when identifying the impulse noise pixel, its value will be replaced by the median value of all the neighbours \cite{Nug2020}.
The difference between the median filtering algorithm and the AMF algorithm is that the filter size of the latter one can be changed based on the characteristics of the input image \cite{Hanu2021}.
As the most frequently adopted technique, AMF is very efficient in restoring image quality, and its implementation flowchart is demonstrated in Figure \ref{fig:amf}.
Apart from annotation and noise removal, images with different scales should also be resized to reach consistency, and examples can be found in \cite{Vad2021, ZhangX22022}.

Compared to the standardized augmentation techniques, the annotations and markers removal techniques are crucially monotonous.
Medical images, especially ultrasound images, are frequently deployed for thyroid cancer detection through CAD systems.
However, ultrasound images are operator-dependent and highly sensitive, resulting in unavoidable speckle noises.
Among all the identified studies, scanty studies proposed noise removal techniques apart from the AMF algorithm; thus, the pre-processing noise removal step is considerably limited.

\subsubsection{Segmentation}

With the objective to detect thyroid cancer through CAD applications, segmentation of medical images is crucial in accurate automatic nodule assessment \cite{Mate2020}.
Image segmentation task allows extracting ROI from the background, mitigating computational resources to diagnose the overall image, and enhancing diagnostic performance by omitting insignificant features.

Medical image segmentation subsumes varied exceptional techniques tailored for semantic entities extraction in the computer vision domain \cite{Ghosh2019}.
For instance, \cite{Prab2018} once evaluated three segmentation algorithms, including active contours without edges (ACWE), graph cut, and pixel-based classifier using thyroid ultrasound images.
Specifically, the ACWE technique initialises an ROI from the input image manually, in which the circled region will be denoted as $1$, and the rest of the ROI will be denoted as $0$ as background.
Then, the ROI will be computed by using the Euclidean distance.
The image force will be calculated through Eq. \ref{eq:force}, in which $I$ represents the original image, $\mu$ denotes the average value inside/outside the contour.
Afterwards, the centre of mass will be calculated for the segmented thyroid nodule; this helps find the potential centre of mass of the nodule in the following image.

\begin{equation}\label{eq:force}
    F_{image} = \int_{inside} (I - \mu_{in})^2 + \int_{outside} (I - \mu_{out})^2
\end{equation}

The second segmentation method \cite{Prab2018} utilised is named ``graph cut'', which was inspired by the ``GrabCut'' technique proposed by \cite{Rother2004}.
Like the ACWE algorithm, the graph cut technique initially requires the manual marking of ROI.
Then, the segmented ROI and the background will form into the Gaussian Mixture Models (GMMs) through clustering; this allows assigning the Gaussian components to the corresponding foreground and background pixels. 
Lastly, the graph cut will find the new foreground and background pixels.

The third method is the pixel-based classifier that users will need to click inside and outside of the ROI, and the features from both regions will be passed into the decision tree for training.
In this case, the extracted features are the coefficient of variation and the mean of the two neighbourhoods.
Thus, the decision tree can automatically learn the foreground and background from those features.
However, this technique requires rigour selection of ROI; otherwise, the segmentation of thyroid nodules can result in erroneous.

Apart from the aforementioned techniques, \cite{Prab2018} also deployed the most classic convolutional neural network (CNN) architecture (i.e., U-Net) for thyroid nodule segmentation.
With the emergence of deep learning, artificial neural networks (ANN), convolutional neural networks (CNN), and recurrent neural networks (RNN) have become backbones for image analysis, pattern recognition, and computer vision tasks.
ANN and CNN are usually applied for classification tasks, whereas RNN is tailored for natural language processing for time-series predictions.
The concept of CNN can be traced back to the $1980s$ \cite{Geron2019}, as it emerged from the study of the brain’s visual cortex and has been widely used for image classification and pattern recognition since then. 
CNN contains three components, including the convolutional operation, pooling operation, and fully connected operations.

{\bf \emph{Convolutional Operation}} In a feed-forward CNN architecture, convolutional layers are used as feature extraction operations \cite{Rawat2017}. 
The convolutional layer breaks the input image into different local regions such as edges and textures. 
The first convolutional layer focuses on extracting low-level features and then assembling those features into a high-level abstract feature map accordingly. 
Noteworthy, inputs were convolved with weights to form a feature map, and all neurons in the same feature map have constraint equal weights \cite{Rawat2017}.
Accordingly, the feature map was calculated in Eq. \ref{eq:fm}, where $Y_k$ is the feature map for the $k_th$ convolution, $f(.)$ is the activation operation, $w_k$ is the weight for neurons in the $k_th$ step, and $X$ is the input image.
Conventionally, the sigmoid function was used as the activation operation, yet today, the ReLU activation is more common for producing non-linearity \cite{Lecun2015}. 
The ReLU activation formula is presented in Eq. \ref{eq:relu} \cite{Shang2016}.

\begin{equation}\label{eq:fm}
    Y_K = f(w_k \times X)
\end{equation}

\begin{equation}\label{eq:relu}
    ReLU(x)=\left\{
    \begin{array}{ll}
      x, & \mbox{if $x\ge0$} \\
      0, & \mbox{otherwise}
    \end{array}
  \right.
\end{equation}

\begin{figure}[h]
    \centering
    \includegraphics[width=0.7\columnwidth]{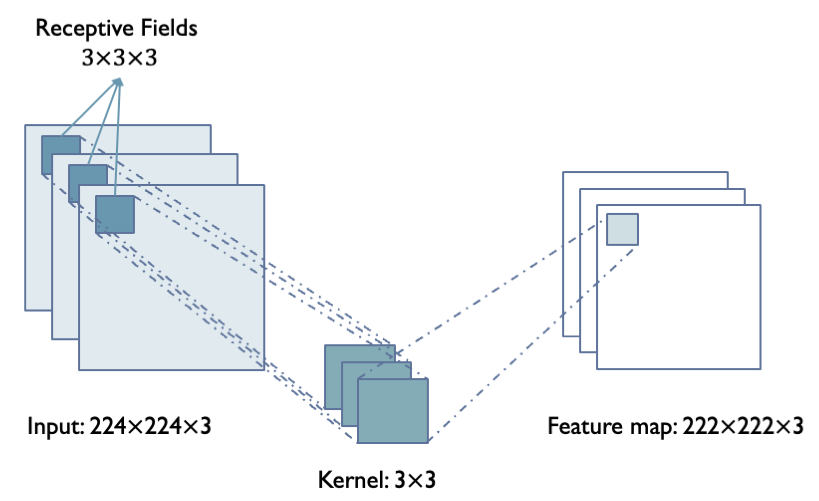}
    \caption{Convolutional operation}
    \label{fig:co}
\end{figure}

The convolutional operation has three hyper-parameters: filter, padding, and stride.
Kernel size, also known as filter size, determining the dimensions of the sliding window over the input image, has been proven to be influential to the performance of CNN. 
The typical kernel sizes for CNN are $3\times3$, $5\times5$, and $7\times7$, sometimes the $1\times1$ kernel size will be used to maintain the feature map size while adjusting the channel numbers. 
Padding is for conserving input images at the borders of activation maps, and usually, zero-padding is adopted for saving computational resources. 
The stride is the steps taken by the sliding window in which the kernel size decided to shift over at a time. 
Figure \ref{fig:co} presents a convolutional operation where the input image is with width and height in RGB color as $224\times224\times3$, when traveling through the kernel size of $3\times3$, the receptive fields are the reflections in the input image with the same size of filters, then the output feature map size was calculated using Eq. \ref{eq:fs}.
Specifically, $n^h$, $n^w$, and $n^c$ are the input image size, $k$ is the kernel size, and $s$ is the stride steps.

\begin{equation}\label{eq:fs}
    f = (\frac{\mathrm{n^h} - k}s + 1) \times (\frac{\mathrm{n^w} - k}s + 1) \times n^c
\end{equation}

\begin{figure}[h]
    \centering
    \includegraphics[width=0.65\columnwidth]{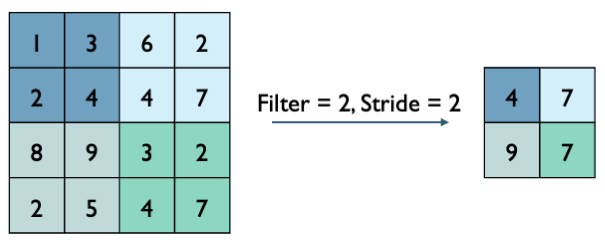}
    \caption{Max-pooling operation}
    \label{fig:pool}
\end{figure}

{\bf \emph{Pooling Operation}} Pooling layers are essentially used to reduce the size of feature maps generated by convolutional operations, thus reducing the computational load. 
In addition, pooling layers only have kernel size and stride step, and no padding parameters are for pooling layers. 
Pooling neurons also have no weights; it aggregates the input using average and max operations functions.  
Figure \ref{fig:pool} presents the max-pooling operation.

{\bf \emph{Fully Connected Layer}} A fully connected layer converts the output from convolutional and pooling operations into a vector representing the probability that a particular feature belongs to a class. 
Moreover, the softmax function is usually used in a fully connected layer to make classifications or predictions for the input image, determining the class of the object.

\begin{figure}[h]
    \centering
    \includegraphics[width=0.9\columnwidth]{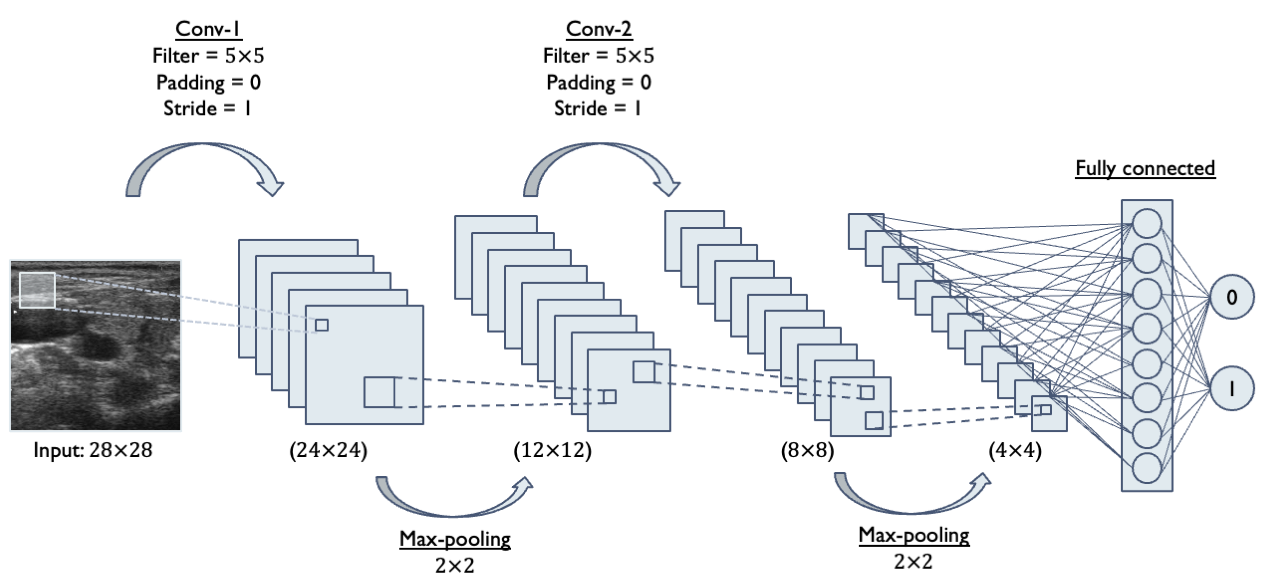}
    \caption{Standardised CNN structure}
    \label{fig:cnn}
\end{figure}

In general, the first CNN architecture was Le-Net introduced in $1998$ \cite{Lenet1998}, since then, AlexNet \cite{AlexNet2012} and VGG \cite{VGG2014} were developed for classification purposes.
These CNN models are in standardized architectures (Figure \ref{fig:cnn}) that stacks several convolution operations, following max-pooling or average-pooling layers, and ended with fully connected layers.

\begin{figure}[h]
    \centering
    \includegraphics[width=0.8\columnwidth]{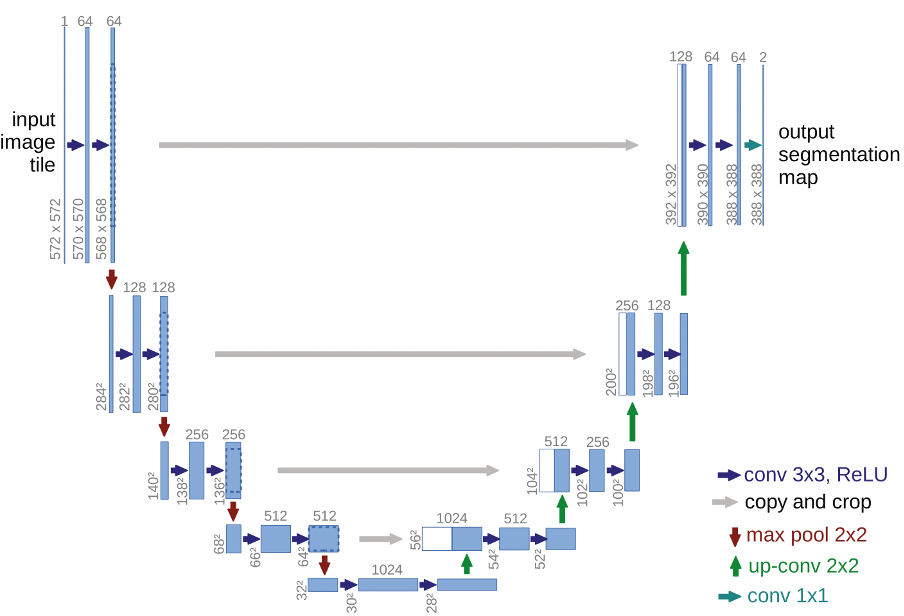}
    \caption{U-Net architecture \cite{Olaf2015}}
    \label{fig:unet}
\end{figure}

With the increasing power of computational resources, more advanced CNN architectures were designed for image classification or segmentation tasks, such as ResNet \cite{He2016} and Inception \cite{Szeg2015}.
Among them, U-Net is a well-known segmentation CNN model, which consists of down-sampling, up-sampling, and skip connection modules, aiming for biomedical image segmentation tasks \cite{Olaf2015} (architecture can be found in Figure \ref{fig:unet}.
The down-sampling layers of U-Net utilise convolutional operations to extract features from the input image.
The up-sampling layers restore the extracted features using the down-sampled latent information.
The skip connections feed the down-sampling feature maps to the corresponding up-sampling feature maps; in the meantime, crop the image from down-sampling to up-sampling to ensure the size is consistent.
Besides \cite{Prab2018}, other researchers also deployed the U-Net architecture for thyroid nodule segmentation \cite{Zhou2018, Mate2020, Sten2021, Dau2021}.

Although the U-Net architecture is the most classic segmentation baseline, many researchers proposed varied models built upon it.
For instance, \cite{Ding2019} modified the model to embed residual and attention blocks to the original U-Net and called ReAgU-Net.
The model presented an increased dice similarity coefficient (DSC) score of $0.869$ compared to the U-Net of $0.820$ on thyroid nodule segmentation.
The DSC measure is a commonly used evaluation metric for segmentation tasks, and it is calculated through Eq. \ref{eq:dsc}, in which TP stands for true positive, FP stands for false positive, and FN stands for false negative.

\begin{equation}\label{eq:dsc}
    DSC = \frac{2 \times TP}{2 \times TP + FP + FN}
\end{equation}

\begin{figure}[h]
    \centering
    \includegraphics[width=0.9\columnwidth]{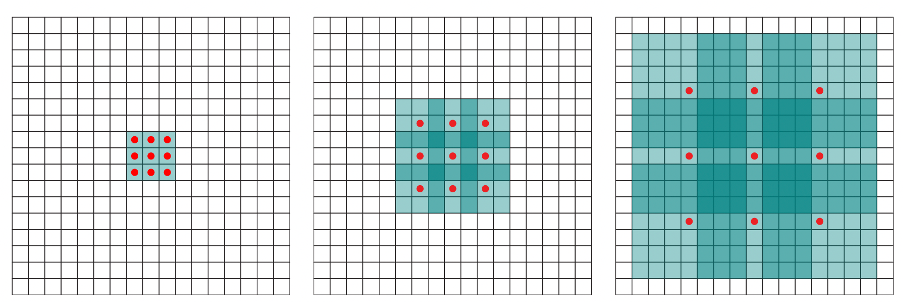}
    \caption{Dilated convolution \cite{Yu2016}}
    \label{fig:dc}
\end{figure}

Similarly, \cite{Xiu2020} proposed attention-based U-Net.
\cite{Nug2021} deployed Res-U-Net proposed by \cite{Cao2020} to segment thyroid nodules on ultrasound images.
\cite{Qing2022} built a dual-route mirroring U-Net called ``DMU-Net''.
\cite{Yiwen2020} proposed a cascade U-Net for thyroid nodule segmentation and classification.
\cite{Shah2021} introduced residual dilated U-Net (resDUnet) for thyroid nodule segmentation, where \cite{Kumar2020} also deployed dilation in the same task.
The dilated convolution is referred to ``convolution with a dilated filter'' \cite{Yu2016}.
In particular, the dilation in convolutional operation supports the exponential expansion of the receptive field without loss of resolution.
Figure \ref{fig:dc} demonstrates the dilated convolution operation.
The first subset is when the input image produces a standard $3\times3$ convolution with dilation equal to $1$.
The second image is produced using $3\times3$ kernel and $2\times2$ dilation rate, generating the receptive field size of 7×7.
The third image is produced when utilising a $4\times4$ dilated rate, generating a receptive filed size of $15\times15$. 
Through the use of dilated convolutions, the number of parameters is identical, whereas the receptive field scale grows exponentially, allowing more features to be captured. 

Owning to performance improvements in deep learning techniques, there has been a concurrent rise in researchers shifting their focus from simply CAD application to designing more exquisite and sophisticated models or modules for thyroid nodule segmentation. 
\cite{liX2018} applied a fully convolutional network (FCN) for this task, in which the model only performs convolution, pooling, and up-sampling.
Likewise, \cite{Hai2021} proposed an encoder and decoder-based FCN model for thyroid nodule segmentation and achieved $81.26\%$ DSC outperforming U-Net.
\cite{Zhou2021} also deployed encoder-decoder structure on thyroid ultrasound images.
Those segmentation-related studies harnessed the power of deep learning algorithms dedicated to detecting thyroid ROI more efficiently and precisely, leading to enhanced classification accuracy.

\subsubsection{Feature Extraction}

Thyroid cancer diagnosis follows a standard clinical procedure, including medical imaging for detecting abnormality inside thyroid glands, FNAC, and excisional biopsy if necessary for pathological examinations. 
For medical imaging analysis, various types of medical image modalities can be deployed, such as ultrasound images, CT scans, magnetic resonance imaging (MRI), radio-iodine scintigraphy, and positron emission tomography (PET) scans. 
Among those medical image modalities, ultrasonography is the most common medical image modality for thyroid cancer detection.
Accordingly, the features manually extracted from ultrasound images are called ``hand-crafted features''. 
Most existing studies consider geometric, morphological, and texture hand-crafted features from ultrasound images in a CAD setting.
Geometric features are the information used to evaluate an object with geometric elements to describe the shape of irregularity \cite{Gomes2020}.
Morphological features are information about lesions' morphological characteristics \cite{Yang2020}.
Additionally, texture features are represented by an image's contrast \cite{Yang2020}.
\cite{Gomes2020} extracted geometric and morphological features from the open-access Digital Database of Thyroid Ultrasound Images (DDTI). 
By augmenting the image set, they have extracted $19$ geometric (e.g., solidity, rectangularity, Orientation, roundness, centroid, etc.) and $8$ morphological features (e.g., area, perimeter, area perimeter ratio, etc.) from $3,188$ ultrasound images.
With the extracted features, the classifier reached an accuracy of $99.33\%$ for detecting malignant thyroid nodules.
Similarly, \cite{Nug2020} once deployed $9$ geometric features (e.g., circularity, compactness, convexity, solidity, etc.) from $165$ ultrasound images.
Through a standardised pre-processing, segmentation of ROI, and ultrasound feature extraction, the neural network reached an accuracy of $0.9479$ classification performance.

Although geometric features are ordinary and sufficient for thyroid nodule detection, integrating various feature sets will undoubtedly enhance the cancer diagnostic rate.
Therefore, besides deploying a solitary type of feature set from ultrasound images, \cite{Yang2020} combined textual and corresponding morphological features (e.g., solidity, centroid, bounding box, etc.) for thyroid cancer detection, and their work obtained a diagnostic accuracy of $89.13\%$.
With all the feature extraction-related studies, fair to moderate agreements were reported that geometric and morphological features are beneficial for establishing thyroid cancer detection CAD systems.
Table \ref{tab:fe} summarises the frequently utilised ultrasound features for thyroid cancer detection.  

\begin{table}[h]
    \centering
    {\bf \caption{Commonly Extracted Ultrasound Features}}
    \begin{tabular}{p{5cm}p{10cm}}
        \hline
        {\bf Feature} & {\bf Description} \\
        \hline
        Area & Area occupied by the object \\
        Solidity & Extent of a shape’s convexity \\
        Centroid & Center of the object \\
        Perimeter & Number of pixels of the object border \\
        Convexity & Ratio of the convex hull to the contour of the object \\
        Tortuosity & Comparison between an object's major axis length and perimeter \\
        Elongation & Ratio between length and width of an object's bounding box \\
        Roundness & Likelihood of an object being round \\
        Filled Area & Total number of pixels inside the bounding box \\
        Orientation & Direct or a position of an object \\
        Eccentricity & Ratio between major and minor axis length \\
        Convex Hull & A convex structure within a detected object \\
        Convex Area & Area surrounded by the convex hull \\
        Aspect Ratio & Ratio between object depth and width \\
        Compactness & Irregularity of an object \\
        Bounding Box & Bounding shape of an object \\
        Rectangularity & Likelihood of an object being rectangular \\
        Elliptic Variance & Comparison of an object shape to an ellipse \\
        Circular Variance & Comparison of an object shape to a circle \\
        Convex Perimeter & Perimeter of the convex hull given an object \\
        Major/Minor Axis Length & Lines lengths through the center of an object \\
        \hline
    \end{tabular}
    \label{tab:fe}
\end{table}

Besides the geometric and morphological features, radiomics analysis is also used commonly as a thyroid ultrasound feature extraction technique, and examples can be found in \cite{Fu2020, Xav2022}.
Radiomics analysis was introduced by \cite{Phil2012}.
It sought to extract critical features from biomedical images for detecting a particular disease by converting medical images into quantitative numbers \cite{Ran2021}.
Specifically, the work conducted by \cite{Fu2020} has extracted $1,079$ radiomics features from ultrasound images to detect lymph metastasis for thyroid cancer patients.
Similarly, \cite{Xav2022} extracted radiomics features from $302$ thyroid nodules.

Apart from using ultrasound images, a few studies proposed other medical image modalities for feature extraction in thyroid cancer detection.
For instance, \cite{Ran2021, Wen2020} both applied radiomics analysis on magnetic reasoning images (MRI).
\cite{Wei2019, YanZ2020} adopted computed tomography (CT) scans for radiomics analysis.
Additionally, \cite{Wu2021} adopted morphological features from CT scans, and their work reached a classification accuracy of $77.7\%$ for detecting PTC.

Ultrasound images have high dimensionality, and extracted features are usually robust yet complicated.
In this regard, it is recommended to reduce the redundant and irrelevant features for enhanced classification performance.
Therefore, feature selection algorithms are generally inducted for recruiting vital features to achieve accurate disease detection.
\cite{LiuTJ2017} once compared five feature selection algorithms with ultrasound image features, including three filter-based and two embedded-based techniques.
Feature selection algorithms can be fallen into three categories, including filter-based, wrapper-based, and embedded-based \cite{LiuTJ2017}.
Filter-based feature selection algorithms consist of two stages, including ranking the features and making classifications with the highest-ranked features, and classic filter-based feature selection techniques are Chi-square, Pearson's coefficient, and Mutual information \cite{Jiliang2014}.
Nevertheless, the most significant limitation of filter-based algorithms is that they can only select features independently, therefore, overlooking the relevance or associations among varied features.
On the contrary, wrapper-based methods select features iteratively by searching through a subset of feature combinations for classification evaluation, and traditional wrapper-based techniques are recursive feature elimination and genetic algorithms \cite{Jiliang2014}.
Embedded-based techniques usually select features directly from the modelling process, typical example is decision tree \cite{LiuTJ2017}.
Based on their work, the utilised filter-based algorithms were Maximum Class Separation Distance, Fisher Scores, and Relief.

Maximum Class Separation Distance is a feature selection method sorting the differences between benign and malignant ultrasound image features, and they selected features with the largest differences \cite{LiuTJ2017}.
The differences can be calculated through Eq. \ref{eq:mcs} where $N_B$ is the number of benign ultrasound images, $N_M$ is the number of malignant ultrasound images, and $f$ is the feature vector for the $i$th image.

\begin{equation}\label{eq:mcs}
    Diff = |\frac{\sum_{i=1}^{N_B}f_i}{N_B} - \frac{\sum_{i=1}^{N_M}f_i}{N_M}|
\end{equation}

Fisher score was introduced by \cite{Jaak1998} to derive a kernel function based on generated probability model so that it can select features with higher discriminate information \cite{LiuTJ2017}, and it is calculated through Eq. \ref{eq:fisher}.
Specifically, $C$ is the number of classes, $N_k$ is the number of $k$th class, $\mu_{jk}$ is the mean for $j$th feature in $k$th class,
and $\sigma_{jk}$ is the  standard deviation of the $j$th feature in $k$th class.

\begin{equation}\label{eq:fisher}
    Fisher = \sum \nolimits_{k=1}^{C}N_{k}(\mu_{jk}-\mu_{j})^{2}/\sum\nolimits_{k=1}^{C}N_{k}\sigma_{j}^{2}
\end{equation}

Relief is the method proposed by \cite{Ken1992}, the Relief method has a high tolerance for noises, and it can select features even though they are interactive.
The Relief algorithm randomly selects a sample and its nearest neighbour in the same class and updates their weights constantly.
In work \cite{LiuTJ2017}, they have selected features from $200$ samples with $5$ nearest neighbours based on the highest feature weights.

With the two wrapper-based feature selection algorithms, \cite{LiuTJ2017} also evaluated their performance.
Penalized Least-Squares Mutual Information aims to find a sparse weight through maximizing the squared-loss variance of mutual information\cite{LiuTJ2017}.
The Hilbert-Schmidt Independence Criterion LASSO (HSIC LASSO) is proposed by \cite{Mako2014} to capture non-linear feature dependency.
Based on the analysis conduced by \cite{LiuTJ2017}, when $90\%$  of classification accuracy obtained, the minimum number of feature sets were used through the HSIC LASSO technique with only $59$ features involved.
The Max Distance however acquired $1400$ feature sets for the classification task.

\begin{figure}[h]
    \centering
    \includegraphics[width=0.85\columnwidth]{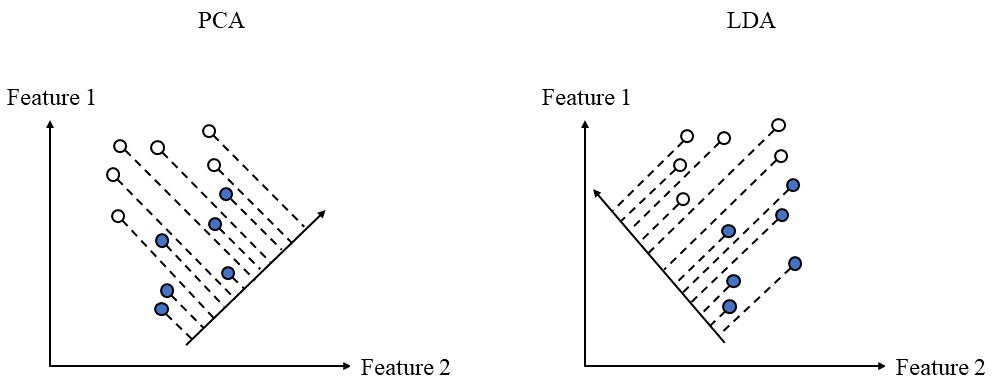}
    \caption{PCA and LDA comparison}
    \label{fig:pca}
\end{figure}

Apart from those, \cite{Hat2020} once adopted the principal component analysis (PCA) technique for selecting features on thyroid ultrasound images. 
PCA is introduced by \cite{Pearson1901} to reduce feature dimensionality, ensure explicit interpretability, and minimize information loss.
PCA analyzes observations that have dependent and inter-correlated variables by extracting information from data and forming a new set of orthogonal variables components. 
It also identifies the patterns in observations and highlights their similarities by calculating the mean centring for each dimension of observation, then following the eigenvalue decomposition of the data matrix \cite{Abdi2010}, and PCA is frequently adopted in thyroid disease detection \cite{Li2012, Zhang2016}.
\cite{Li2012} also adopted PCA as feature extraction to distinguish instances among ``normal'', ``hypothyroidism'', and ``hyperthyroidism''.
\cite{Zhang2016} applied PCA with an ensemble method for thyroid disease detection.

Besides PCA, linear discriminant analysis (LDA) is also another feature selection algorithm used quite commonly.
LDA reduces feature dimensions yet focus more on separating the known classes to the maximum extend. 
PCA aims to maximize the variance of the input features, while LDA focuses on maximizing the distance between different categories. 
Figure \ref{fig:pca} presents their difference. 

\begin{table}[h]
    \centering
    {\bf \caption{Ultrasound Features Selected for Thyroid Cancer Detection}}
    \begin{tabular}{p{5cm}p{10cm}}
        \hline
        {\bf Reference} & {\bf Features Used} \\
        \hline
        \cite{Lim2008} & Intercept, size, shape, margin, echogenicity, cystic change, micro-calcification, halo \\
        \cite{Yueyi2008} & Micro-calcification, shape, margin, capsular invasion, architecture, echogenicity, ring down artifact, vascularity \\
        \cite{Chang2010} & Gray level co-occurrence matrix, statistical feature matrix, gray level run-length matrix, law’s texture energy measure, neighboring gray level dependence matrix, wavelet features, Fourier features \\
        \cite{Zhu2013} & Shape, margin, echogenicity, internal composition, presence of calcification, peripheral halo, vascularity \\
        \cite{Ach2014} & Shape, echogenicity, calcification, echo texture, margin, capsule invasion, halo \\
        \cite{KimE2016} & Size, area, shape, color, texture of regions, histogram of oriented gradients, co-occurrence gray level matrices, chromatin density \\
        \cite{Chang2016} & Histogram, intensity differences, elliptical fit, gray-level co-occurrence matrix, gray-level run-length matrix \\
        \cite{WuH2016} & Location, position, shape, margin, internal contents, echogenicity, calcification, echo-genic foci in solid portion, halo, infiltration and extracapsular invasion, increased intra-nodular vascularity, abnormal lymphadenopathy, multifocal \\
        \cite{LiuT2017} & Gray level co-occurrence matrix, Local binary patterns, Histogram of oriented gradient, Scale-invariant feature transform, vector of locally aggregated descriptors \\
        \cite{Ko2019} & Margin, internal content, anteroposterior dimension-to-transverse dimension ratio, microcalcifications \\
        \cite{Tian2019} & Size, margin, shape, aspect ratio, composition, calcification \\
        \cite{Liang2020} & Size, morphology, location, echo, margin, boundary, surrounding tissue, posterior echo, calcification \\
        \cite{HaoW2020} & Aggressive histology subtype, vascular tumor capsular invasion, extra-thyroidal extension, regional metastases, and distant metastases \\
        \cite{Gitto2019} & Composition, echogenicity, orientation, margin, shape, spongiform, calcification, elasticity, vascularity \\
        \cite{Yoo2018, Jeo2019} & Composition, shape, margin, orientation, echogenicity, spongiform status \\
        \cite{Car2018, Guan2019, Shin2020} & Composition, echogenicity, calcification, margin, shape \\
        \cite{Xia2017, Ouyang2019, ZhangB2019} & Size, margins, shape, aspect ratio, capsule, hypo-echoic halo, internal composition, echogenicity, calcification pattern, vascularity, and cervical lymph node status \\
        \hline
    \end{tabular}
    \label{tab:sf}
\end{table}

Based on the identified studies, researchers tend to apply various feature selection algorithms to establish thyroid cancer diagnostic systems, and the commonly selected features from ultrasound images are summarized in Table \ref{tab:sf}. 
The most commonly used features are nodular size, shape, margin, composition, and calcification presence. 
Those knowledge extracted from medical images will then be fed into machine learning or deep learning classifiers for evaluate the thyroid status for disease detection.

\subsubsection{Classification}
With the development of electronic computers, algorithms were developed rapidly in the $1950$s. 
Three basic machine learning algorithms emerged then, including symbolic learning, statistical learning, and neural networks. 
Those three branches grew more advanced and become the well-known classifiers today ``decision trees, K-nearest neighbours (KNN), and multi-layer feed-forward neural networks'' \cite{Konon2001}.
In this regard, various CAD systems built upon those classifiers, such as support vector machines (SVM) to diagnose breast cancer \cite{Arafa2019}, Naive Bayes (NB) to detect brain tumour \cite{Sharma2014}, multi-layer perceptron to diagnose liver cancer \cite{Naeem2020}, to name a few. 
Generally, the essence of machine learning is to train computers to learn from data through statistical analysis. 
Deep learning is emerged as a subset of machine learning and has become an intense tool for computer vision tasks.
Recently, deep neural networks have been used frequently in CAD design for helping to make diagnostic decisions and have shown satisfying performance, such as detecting diabetic retinopathy \cite{Zeng2019}, Covid-$19$ detection \cite{Wang2020, Apos2020}, and malaria diagnosis \cite{Liang2016}. 
Studies using machine learning-based CAD application to detect thyroid cancer is also abundant.

\cite{Liu2008} once implemented the Naïve Bayes (NB) classifier on $41$ thyroid nodules ($21$ benign and $20$ malignant) from $37$ patients and compared with two experienced radiologists. 
The results obtained by the NB algorithm outperformed the radiologists and reached an area under the curve (AUC) of $0.851$. 
Similarly, \cite{Singh2012, Xia2017, Ouyang2019} acquired self-obtained ultrasound features for making classifications and obtained comparable results to experienced radiologists. 

With varied image characteristics (i.e., image quality, image size, image scale), researchers usually spend a significant amount of time evaluating heterogeneous feature extraction algorithms for classification purposes. 
To mitigate the manual process during feature selection, deep learning techniques emerged to help make the diagnosis much more efficiently by extracting and selecting features automatically from input images and making classification decisions correspondingly. 
To this extent, researchers can shift their focus to designing a more accurate and adapted classifier for disease detection. 

As introduced earlier, deep learning is a subset of machine learning, and included but not limit to ANN, CNN, and RNN models.
In some earliest works, ANN has been adopted relatively often using ultrasound images to make a thyroid cancer diagnosis. 
According to \cite{Zhu2013}, $689$ thyroid nodules were examined using ANN and reached an accuracy of $83.1\%$. 
Similarly, \cite{Shin2020} compared SVM classifier with ANN through $348$ thyroid nodules, and the diagnostic accuracy rates are $69\%$ and $74\%$ respectively. 
More recently, CNN has been adopted more frequently regarding detecting thyroid cancer through ultrasound images.

CNN models play vital roles in the computer vision domain; not only they can be used for segmentation tasks, but also for localisation and classification tasks.
Thyroid cancer diagnosis through CNN is natural and intrinsic.
For instance, \cite{Li2019} acquired $131,731$ ultrasound images for cross-institutional analysis through the CNN model.
Their work demonstrated an accuracy of $0.889$, $0.856$, and $0.915$ for the Tianjin, Jilin, and Weihai cohorts.
\cite{Buda2019} acquired ultrasound images from Germany and applied CNN for thyroid cancer detection, and their work obtained an accuracy of $0.78$, which was lower than experienced radiologists.
\cite{Yic2021} proposed a generic eight-layer CNN model for classifying thyroid and breast lesions, and with thyroid cancer detection, the accuracy rate was $86.5\%$ on $719$ ultrasound images.
\cite{NugH2021} used the Inception model to detect thyroid cancer and reached an accuracy of $87.2\%$ with the DDTI ultrasound database.
\cite{Chan2021} once compared VGG$19$, ResNet$101$, and InceptionV$3$ models on $812$ ultrasound images, and the best-performing model was ResNet with $77.6\%$ accuracy rate reached.

\begin{figure}[h]
    \centering
    \includegraphics[width=0.9\columnwidth]{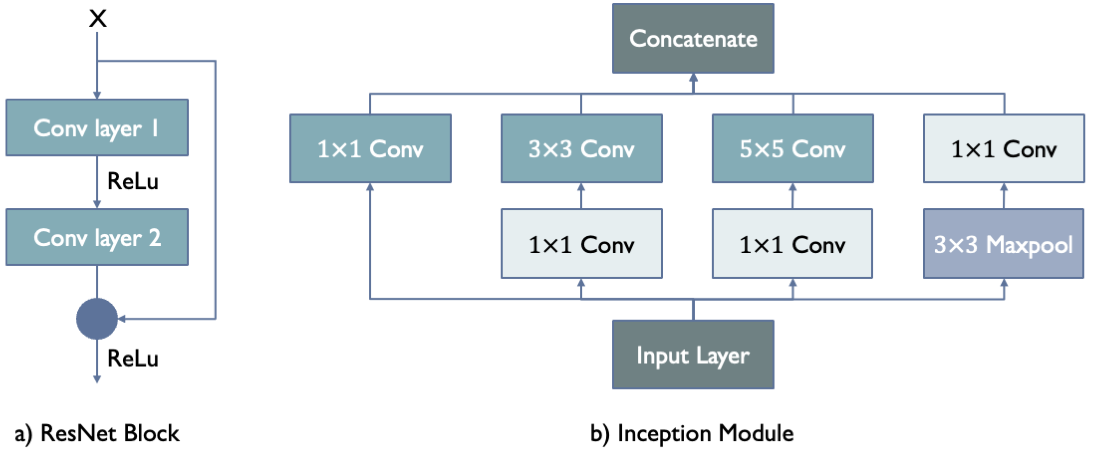}
    \caption{ResNet and Inception Blocks}
    \label{fig:resincp}
\end{figure}

VGG models are generic and adaptive as the convolution operations and the fully connected layers can be adjusted for a specific task. 
The commonly used architectures for VGG are VGG-F, VGG$11$, VGG$16$, and VGG$19$.
Theoretically, the more convolutional operations we stack to the CNN, the better the classification results will be; nevertheless, this is not the real-world case.
The reality is that when more convolutions are stacked, the model will likely get a gradient explosion.
In this regard, the ResNet model was proposed to address the issue as it can generate very deep CNNs, avoiding aggregating parameters that take exponentially increased computational resources \cite{He2016}.
More importantly, ResNet produces better classification results than conventional CNN stacking architectures as it can learn residuals through layers.
The Inception model was proposed by \cite{Szeg2015}, allowing for stacking multiple Inception modules to generate more accurate results.
Each Inception module concatenates feature maps generated through different kernel sizes with different information learned through the varied size of receptive fields.
Figure \ref{fig:resincp} demonstrates a ResNet block and an Inception module.

The use of CNN architecture demonstrates varied performance following the different image analysis steps with heterogeneous image quality.
Under most situations, the proposed models are limited in generalising to different datasets, and there is a possible over-fitting concern for most previous works.
In this regard, transfer learning has been applied quite often to mitigate the over-fitting phenomenon.
Transfer learning uses parameters learnt from pre-trained neural networks and applies those ``gained knowledge'' to new tasks by freezing the previous layers and making changes to the last few layers. 
For example, \cite{Ma2017} proposed the integration of two pre-trained CNNs through transfer learning, where the shallower network was used for learning low-level features, and the other deeper network was used to learn high-level abstract features. 
Then the two learned feature maps from two CNNs were fused as an input into a softmax layer to diagnose malignant thyroid nodules, resulting in diagnostic accuracy of $83.02\%$. 
Furthermore, \cite{Chi2017} fine-tuned an Inception model and tested it on two ultrasound databases. 
For the DDTI set, the accuracy was $98.29\%$, and for the private data set, the accuracy was $96.34\%$. 
A moderate consensus was made that CNN applications on ultrasound images for detecting thyroid cancer are efficient and accurate \cite{Ko2019, Guan2019, Liang2020, Olu2022}.

\begin{figure}[h]
    \centering
    \includegraphics[width=0.65\columnwidth]{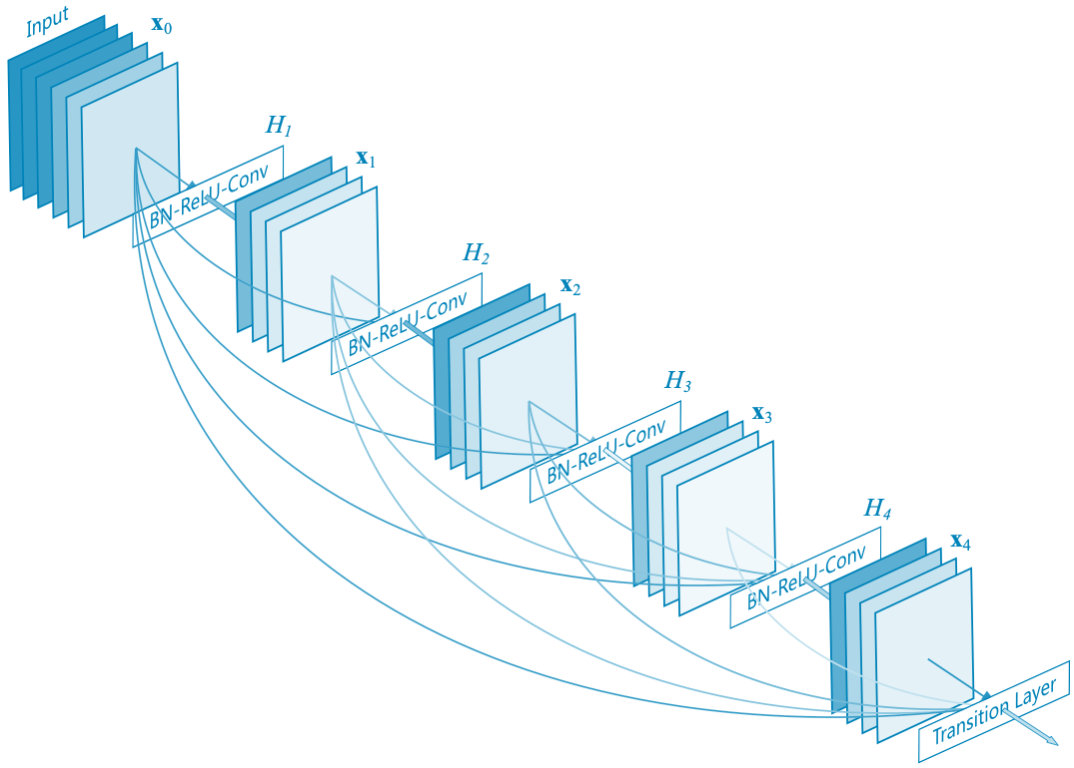}
    \caption{Dense Blocks}
    \label{fig:dense}
\end{figure}

There has been a concurrent rise in applying other image modalities for thyroid cancer detection rather than ultrasound.
For example, \cite{Wang2019} adopted VGG-$19$ and InceptionResNetV$2$ to multi-classify thyroid nodules into seven classes through histopathology images, including normal tissues, adenoma, goiter, papillary cancerous nodule, follicular cancerous nodule, medullary cancerous nodule, and anaplastic cancerous nodule. 
The results suggest that VGG-$19$ yields better average accuracy than InceptionResNetV$2$ that is $97.34\%$ and $94.42\%$ respectively. 
Similarly, \cite{Dov2019} adopted the Multiple-instance Learning (MIL) approach on segmented whole-slide images to predict malignancy of thyroid tissues. 
\cite{Vijaya2020} trained ResNet$50$ and DenseNet$121$ models through transfer learning with histopathology images and reached an accuracy of $100\%$.
\cite{Chan2020} acquired FNAC cytological images to detect medullary thyroid cancer, and the CNN model reached an approximated classification accuracy of $99.00\%$.
The DenseNet was proposed by \cite{Gao2017} where it contains three dense blocks.
Each layer from a dense module connects to all its preceding layers as the input (Figure \ref{fig:dense}).

Apart from those, \cite{Lee2019} adopted eight CNNs on CT scans to differentiate thyroid cancerous metastasis, including DenseNet$121$, DenseNet$169$, InceptionResNetV$2$, InceptionV$3$, ResNet, VGG$16$, VGG$19$, Xception. 
Based on their comparison, the best AUC was obtained by InceptionV$3$ and ResNet, around $0.95$.
\cite{ZhangX2022, ZhangX22022} compared ultrasound and CT modalities on CNN performance and found that ultrasound has a slightly better result than CT.
\cite{Rong2018} utilised MRI on multi-modality CNN for classification and reached a diagnostic accuracy of $0.82$ on $45$ images.
\cite{Naglah2021} constructed a multi-input CNN for thyroid cancer diagnosis from $49$ patients who underwent MRI tests, and their model reached an accuracy of $0.88$.
We found that ultrasound seems to lose its dominance in thyroid cancer detection; more medical image modalities were analysed through the use of CNN, including but not limited to CT scans \cite{Zhao2021}, hyperspectral imaging \cite{Hal2017}, and SPECT images \cite{Ma2019}.

\subsubsection{Thyroid Cancer CAD Literature Summary}

Machine learning-based CAD decision support systems for thyroid cancer appear to present a varied performance from $70\%$ to $90\%$; the main underlying reason might be aroused from acquired data sets qualities. 
Medical data sets are challenging to be acquired as they are correlated with ethical and privacy concerns.
With machine learning, it is of primary demand to acquire extensive high-quality data for training unbiased models. 
In the real-world setting, high-quality medical data sets are scarce, much less is to say the expensive labour sources of data labelling and annotations process, which aggravates the challenges of data acquisition. 
Under this scenario, the open-access datasets are valuable for researchers to evaluate their CAD systems and provide baseline comparisons.
Table \ref{tab:data} demonstrates the open-access datasets for thyroid disease detection tasks frequently used in existing studies.

\begin{table}[h]
    \centering
    {\bf \caption{Open-access datasets for thyroid disease detection}}
    \begin{tabular}{p{9cm}ll}
        \hline
        {\bf Data Source} & {\bf Data type} & {\bf Sample Size} \\
        \hline
        Gene Expression Omnibus repository & Numerical & $25$ samples \\
        UCI Machine Learning Repository & Numerical, categorical & $7,200$ instances \\
        Knowledge extraction based on evolutionary learning	& Numerical, categorical	& $7,200$ instances \\
        Kaggle database	& Numerical, categorical & $3,772$ instances \\
        Digital database of thyroid ultrasound images (DDTI) & Ultrasound images & $400$ instances \\
        Wilmington endocrinology database &	Ultrasound images &	$59$ images \\
        Papanicolaou Society of Cytopathology &	FNAC images & $110$ images \\
        National Cancer Institute TCGA & Whole-slide image & $507$ instances \\
        \hline
    \end{tabular}
    \label{tab:data}
\end{table}

Since machine learning approaches expect scholars to compare their proposed works with existing benchmarks, open-access databases are vital. 
With the limited number of sample instances in self-acquired datasets, not only the use of open-access datasets is essential, but also data augmentation is critical.
Data augmentation methods for medical images are in the cookie-cutter stereotype, such as flipping or rotating.
However, annotations on each medical image suggest the nodule location. 
Changing the image's direction will correspondingly affect the nodule position; this, in turn, may influence the model's performance.
Additionally, data augmentation methods were usually applied to medical images, whereas studies utilising clinical features (e.g., morphological or texture information) scarcely deployed such augmentation steps.
Therefore, standardised data augmentation techniques may work well on natural images, whereas they are not appropriate for medical images; future research needs to investigate more solid and suitable augmentation techniques for medical images and clinical features.

The CAD process is tedious and requires sufficient training, evaluation, and selection of the most appropriate solution.
In the clinical domain, efficiency is just as imperative as accuracy; thus, removing unnecessary steps is always encouraged.
In the segmentation tasks, removing annotations and segmenting ROI can be conducted simultaneously using De-noise auto-encoders.
They are efficient in noise removal and have already been applied in the clinical domain for breast lesion detection \cite{Jie2016} and brain lesion detection \cite{Var2017}.
However, the use of this architecture in thyroid cancer detection is still absent.
Future research studies should focus on shortening the CAD diagnostic efficiency without harming its accuracy and propose more effective, accurate, and adaptive models.

Moreover, diagnosis is correlated with precision treatments to achieve a promising prognosis. 
Thyroid cancer has several kinds of subtypes, all resulting in different treatment protocols. 
Therefore, for the sake of customised treatments and a well-established prognosis, the diagnosis should be made more targeted. 
However, existing studies mainly focus on the binary classification to detect benign and malignant thyroid nodules. 
Among all the identified thyroid cancer CAD-related literature studies, only a few have applied multi-class classification tasks to determine thyroid cancer subtypes \cite{Wang2019, ZhangX22022}. 
Hence, studies around the multi-class classification of thyroid cancer are significantly limited. 
More studies should be proposed from the perspective of multi-classifying thyroid cancer subtypes to achieve comparative diagnoses made by experienced clinicians.

Furthermore, most existing studies are dedicated to applying ultrasound images for making diagnoses through machine learning techniques. 
Thyroid disease can be diagnosed through different medical images; relying on a single type of medical image may resist the diagnostic accuracy. 
Therefore, more approaches should be designed using different medical data sets rather than depending on ultrasound images for evaluation.
Incorporating text-based reports, clinician notes, or even varied image modalities is recommended to drive a more comprehensive and accurate diagnostic decision.

Those aforementioned shortages create limitations for applying CAD systems in the clinical field.
There is still a long way for clinics to trust and get comfortable with the CAD models, as they need evidence to show their practicability. 
Bridging the gaps, researchers should establish favourable benchmarks for evaluating their machine learning approaches, and CAD systems should also target dealing with limited-quality medical data sets. 
Besides, evidence to show the feasibility and interpretability of deep learning models is demanding so clinicians can evaluate the techniques and comprehend the decisions the systems make.

\subsection{Applications in Thyroid Cancer Prognosis}

With the progress of CAD developments, a manifesto for promoting an accurate diagnosis has been achieved, whereas precision medicine is a prospective phenomenon to be launched.
Individual patients have varied health conditions or disease stages, and treatment generalisation is not appropriate in this regard.
Precision medicine, on the other hand, is tailored for individual health care on the basis of the target's genes, lifestyle and environment \cite{Rich2016}, maximising the patient's health status after targeted treatments.
In the clinical domain, customised treatment plans contrive to betterment the prognosis for individual patients.
Establishing sagacious decision support systems on treatment protocols enlarges the applications of precision medicine.
More importantly, predicting thyroid cancer patients' death, survival, and recurrent rates will potentially guide them to achieve an optimal prognosis.

\subsubsection{Treatment Decision Support System}

Thyroid cancer risk stratification depends on its subtypes, with papillary kind being the least severe kind and anaplastic being the most severe kind based on their death rate \cite{Melis2020}, and different subtypes undergo varied treatment protocols.
Treatment protocols rely on individual patients' health conditions, such as age, weight, body mass index, pregnancy status, medical history, and medication doses for other diseases.
Sometimes, external factors like seasonal temperature, financial status, and the patient's wills also play significant roles in developing treatment plans \cite{Sch2010}. 
To this extent, precision medicine and treatment have to be achieved for the target patient.
With the increasing instances of thyroid cancer, CAD applications have been generated more often, whereas precision medicine development has been somehow neglected in the past few years.
Therefore, customized treatment protocols should be offered to individual patients based on their expectations to design a thyroid cancer-targeted treatment decision support system.  

Among the identified studies, customised treatment decision support systems-based works are limited.
\cite{Chen2018} once adopted the density-peaked clustering analysis technique for disease symptoms clustering. Meanwhile, they adopted the Apriori algorithm for establishing treatment rules, called the Disease Diagnosis and Treatment Recommendation System. 
In \cite{Chen2018}’s work, they have also applied the Apriori algorithm to detect the association relationships between the symptom clusters and the treatment schemes and have yielded satisfying performance; meanwhile, the system's interface was also designed and tested.

\begin{figure}[h]
    \centering
    \includegraphics[width=0.4\columnwidth]{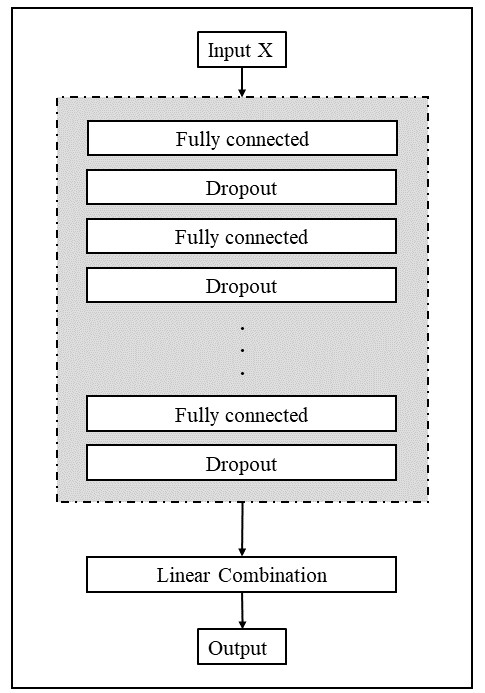}
    \caption{DeepSurv Model \cite{Katz2018}}
    \label{fig:sur}
\end{figure}

\cite{Katz2018} once applied the DeepSurv model to develop personalized treatment protocols for patients with a particular disease. 
The DeepSurv is a feed-forward ``Cox proportional hazards deep neural network'' used to model the interactions between a patient’s covariates and treatment effectiveness \cite{Katz2018}. 
The typical structure of the DeepSurv model is presented in Figure \ref{fig:sur}. 
The input dataset will travel through several stacked, fully connected, and dropout layers and end with a linear function. 
Lastly, the model produces an output of the predicted log-risk function to determine the survival time of a set of patients with personalized treatment protocols established.

\begin{figure}[h]
    \centering
    \includegraphics[width=0.45\columnwidth]{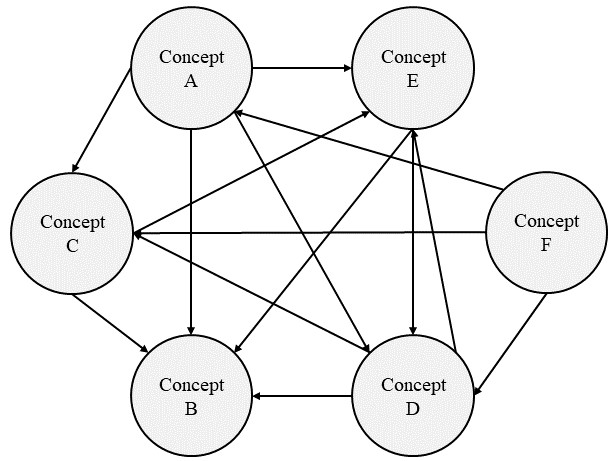}
    \caption{Fuzzy Cognitive Map Architecture}
    \label{fig:fcm}
\end{figure}

Besides the Apriori-based and DeepSurv models, the fuzzy cognitive map approach has been implemented a few times to design a patient-centric treatment decision support system. 
A fuzzy cognitive map (FCM) incorporates ANN and fuzzy logic and shares similar logic with human reasoning and human decision-making \cite{Papa2006}.
Specifically, FCM integrates qualitative and quantitative data, and it looks like a cognitive map consisting of concepts and relationships. 
FCM can model complex systems and is tailored for developing decision-making systems, particularly disease treatment decisions.
FCM consists of concepts and weights. 
The concepts are the representative variables for making treatment decisions, for example, patient age, nodule size, nodule location, metastasis extent, to name a few. 
The directed edges with arrows present the degree of the relationship between interdependent concepts, and they are known as weights.
The fuzzy rules infer with the weights.
Figure \ref{fig:fcm} demonstrates the basic FCM architecture.
The mathematical representations of a FCM can be achieved through Eq. \ref{eq:fcm}.
Specifically, The $A_i^{(k+1)}$ is the value for a concept at simulation step $k+1$. 
$w_{ji}$ is the weight for the interconnection between two concepts.
$N$ is the total number of instances.
$f$ is the sigmoid threshold function. 
And the FCM approach has been applied several times in designing personalised disease treatment recommendation systems \cite{Papa2003, Papa2006, Chry2008}.

\begin{equation}\label{eq:fcm}
    A_i^{(k+1)} = f(A_i^k + \sum_{j=1, j \neq i}^N A_j^k w_{ji})
\end{equation}

\subsubsection{Death, Survival, and Recurrence Prediction }

Several judicious studies proposed utilizing machine learning approaches to evaluate the prognosis of thyroid cancer, including the survival rate, death rate, and recurrence rate of patients diagnosed with thyroid cancer. 
Among the identified papers, ANN has been adopted most often \cite{Cruz2006}.
For instance, \cite{Jaj2014} has applied SEER data on MLP and logistic regression to predict the survival rate for $7,706$ patients with thyroid cancer, resulting in the better performance obtained by the MLP approach.
Similarly, \cite{Kukar1997} once adopted decision trees, regression trees, and the Cerebellar Model Articulation Controller neural network to identify leading factors that might influence the survival of patients with anaplastic thyroid carcinoma. 
Their results showed that age, the growth rate of tumours, metastasis distance, the patient's health condition, performance, and concomitant diseases should all be considered when predicting the individual's survival.
Similarly, \cite{Mourad2020} has applied three ANNs, specifically Multi-layer Perceptron (MLP), to the U.S. SEER-$18$ database, which consists of $25,063$ patients with thyroid cancer, to determine the probability of death rate caused by thyroid cancer. 
Their first MLP has selected input features, including patient gender, age, race, tumour size, primary disease extent, nodular location, and er of positive lymph nodes. 
With a $19$-layer MLP, the ANN would output the patient's status, which is still alive or dead. 
The second MLP only inputs age, primary disease extends, and nodule locations selected by the filter-based feature selection algorithms with $18$ hidden neurons and the same output. 
The third MLP is designed with three input features: tumour size, number of positive lymph nodes, and metastases. 
Moreover, the architecture has $4$ hidden neurons, and the outputs are maintained from the second MLP architecture. 
The best performing model was the first MLP that reached an accuracy of around $94.49\%$.

Researchers also pay close attention to the recurrence rate of thyroid cancer.
For instance, \cite{Park2021} compared five machine learning models to evaluate the recurrence rate of patients with papillary thyroid carcinoma. 
The input features they selected were age, gender, tumour size, tumour multiplicity, lymph nodes metastasis, lymph nodes ratio, extranodal spread, and extrathyroidal extension. 
The selected models were decision tree, random forest, XGBoost, LightGBM, and Stacking. 
The decision tree outperforms the other four and has reached a prediction rate of $95\%$. 
\cite{Yang2019} has applied the Ensemble Algorithm for Clustering Cancer Data (EACCD) algorithm to design a prognosis system for thyroid cancer patients to minimize the probabilities of recurrence. 
Through a set of input features: tumour size, lymph nodes, metastasis, and age, the EACCD algorithm was applied.
It consists of three steps, including defining the initial dissimilarities between classes, applying ensemble learning to obtain the learned dissimilarities, and clustering the combinations of the learned dissimilarities.
\cite{Schneider2013} applied the multivariate logistic regression on $217$ patients' lymph ratios to determine the recurrence of papillary thyroid cancer. 
The results show that the lymph node ratio is a significant factor that correlates with PTC recurrence.

\subsubsection{Thyroid Cancer Prognosis Literature Summary}

Patients diagnosed with thyroid cancer concerns the most about the treatments, recurrence, survival, and death rates.
Based on our analysis, studies on machine learning-based treatment decision support systems are far less than those on CAD systems. 
Among all the identified works, none of them was designed and targeted specifically for thyroid cancer.
Unlike the CAD systems, the treatment decision support systems are much more interpretable and acceptable by clinicians since factors affecting the treatment decision-making are easily qualified and quantified by clinicians. 
Moreover, clinicians are crucially involved in the system's development phase.
However, the works around this area have been considerably ignored. 
With the customised treatment decision support system design, individual patients can obtain optimal prognostications by receiving the most fitted treatment protocols so that the recurrence or even death rates can be dramatically mitigated. 

Based on our analysis, most of the established work was proposed in $2005$.
Nevertheless, today is the artificial intelligence era, and many more advanced machine learning techniques have emerged yet have not been applied in the clinical field for thyroid cancer treatment. 
Hence, we expect a thyroid cancer-specific treatment decision support system to be effectively generated and proposed by adopting current state-of-the-art algorithms by taking into consideration all related factors, such as patient information (i.e., age, gender, BMI, height, weight), medical history, dietary, financial conditions, disease subtypes, illness degree, medication intake, to name a few. 
Through comprehensive factor analysis, an optimal treatment plan can be provided for patients to achieve customized treatments to enhance clinical trust leading to potential adoption.      

Notably, the most challenging issue the researchers face today is the limited data set, where retrospective medical data sets are vital. 
Retrospective data contains patient information from the diagnosis to the prognosis, usually suggested being archived for five to ten years. 
In order to have a valid and feasible algorithm to detect the survival rates (i.e., $5$-year, $10$-year, $20$-year), death rates, and recurrence rates for patients with thyroid disease, it requires a large amount of high-quality retrospective clinical data. 
Therefore, retrospective medical data sets are precious, making the predictions on the prognosis of thyroid disease extremely tricky; that is why this research field expects a data set that is authentic and complete. 

Additionally, machine learning approaches usually take patient age, tumour size, and lymph metastasis factors for making predictions, yet ignore another vital factor, ``time''. 
To develop algorithms that can predict the prognostic status, the ``time'' factor is particularly imperative since clinicians would potentially prepare for the upcoming health changes appearing in the patient. 
Hence, we expect more algorithms like RNN to get involved, which takes time into consideration to make predictions so that patients can better understand their health status to guide them into an optimal health record.

In summary, more advanced machine learning algorithms should be involved in the development of thyroid cancer-specific treatment recommendation systems.
As for the prognostic predictions of thyroid disease, retrospective medical data is challenging to acquire, leading to limited studies in this field. 
Hence, an open-access database will be significantly helpful for promoting the developments in the domain. 
Another worth mentioning point is that time is crucial for survival, death, and recurrent rates prediction, while it has been neglected in previous studies. 
We urgently need the means to see more machine learning approaches that incorporate the time factor into forecasts to help patients prepare so that they can gain a promising prognosis.

\section{Challenges and Recommendations}

Although machine learning and deep learning techniques brought unprecedented advantages to the clinical field, there still exist many challenges and opportunities for researchers to excavate. 
This section highlights the open issues and corresponding future research directions.

Thyroid cancer pathogenesis is always of immediate attention to be discussed and researched in the past decades.
Researchers dedicated to put their effort into identifying risk factors correlated with thyroid cancer through both qualitative and quantitative techniques.
However, many of the factors are still controversial and need further evaluations.
Among the identified studies, the use of data mining techniques to reveal thyroid cancer pathogenesis is still under-researched; future work should investigate more data mining algorithms to address the issue.

With the development of machine learning-based CAD approaches, the diagnosis of thyroid cancer was more accurate with less human involvement. 
Those CAD systems require extensive high-quality data to train a sophisticated neural network. 
However, high-quality medical datasets are rare. 
Currently, most of the well-performing neural networks were pre-trained through transfer learning; this potentially causes an over-fitting issue that might be biased when evaluating the model relying on a single data source. 
Moreover, medical images have different characteristics from natural images on ImageNet \cite{Russ2015}, as medical images are less heterogeneous in structure. 
Therefore, we recommend training a CAD model with a specific thyroid medical image set and evaluating the system using multiple data sources to ensure the fairness and generalisation of the model; this potentially enhances the CAD adoption in the clinical field.

As mentioned earlier, high-quality medical data is rare because standardising the dataset in the pre-processing stage is complicated and time-consuming, and collecting them requires ethical agreements. 
Among those studies that have not utilised open-access data sets, the scale of the acquired private data set is relatively small. 
To address the solutions of the small data set issue, data augmentation is an inevitable procedure. 
Typical data augmentation approaches are synthetic augmentation (to increase data size by generating new data), oversampling, and bootstrapping. 
With medical images, data augmentation can be achieved through cropping, flipping, rotating, and mirroring. 
Therefore, when the input data set is imbalanced or small-scale, data augmentation approaches should be adopted for enhancements.  
Furthermore, we do expect to see more data be donated as open-access for research purposes, and this will undoubtedly offer more opportunities to develop and evaluate more robust systems. 
This can be achieved by offering a dataset where individual researchers can contribute little by little and eventually form into a medical data set containing a tremendous amount of high-quality records and is available to everyone. 

Moreover, existing studies focus their attention on classifying hypothyroid and hyperthyroid, or benign and malignant nodules, rather than detecting thyroid disease subtypes. 
Hence, we expect to see more machine learning approaches proposed for multi-classifying thyroid disease subtypes so that patient-centric treatments can be ensured to reduce morbidity rates.
Additionally, text-based reports and varied medical image modalities complement one another, and they should be incorporated for diagnosis rather than relying on a single type of medical data set that lacks authenticity and limits the accuracy of the proposed work.   

The issue around the machine learning-based treatments for thyroid disease is that most studies applied machine learning approaches to design personalised treatment recommendation systems rather than explicitly built for thyroid disease or based on thyroid-related data set. 
Additionally, the evaluations of those treatment recommendation systems were significantly ignored by researchers. 
We believe there is a significant demand for treatment recommendation systems specifically built for thyroid disease in future works, and more comprehensive factors should be taken into consideration during the developing stage. 
Moreover, the evaluations of those systems must be arisen to enhance the potential adoption in the clinical; this can be achieved by clinicians' involvement to accelerate the design and adoption of such systems.

Furthermore, with the studies on the prognosis of thyroid cancer, the input features for the prognostic, predictive models did not consider ``time'' as an essential attribute, which can immensely concern the performance of any survival, death, or recurrence prediction models.
The time factor is expected to be incorporated in future works to help build an optimal model so that clinicians and patients can be well prepared for the health condition changes appearing in the subsequent period intervals to achieve an ideal prognosis. 

Last but not least, for the diagnostic and prognostic systems with thyroid disease, supervised models were used mainly, while treatment recommendation systems usually establish unsupervised models. 
In reality, unlabeled medical datasets are usually excluded from research, and this leads to huge data waste.
Combining supervised and unsupervised (i.e., semi-supervised) machine learning models might accelerate the utilization of unlabelled data. 
We expect to see more studies to deal with data waste issues in the field by designing potent semi-supervised decision support systems.

\end{document}